\PassOptionsToPackage{unicode}{hyperref}
\PassOptionsToPackage{hyphens}{url}
\PassOptionsToPackage{dvipsnames,svgnames,x11names}{xcolor}
\documentclass[
]{article}
\usepackage{xcolor}
\usepackage{amsmath,amssymb}
\usepackage{iftex}
\ifPDFTeX
  \usepackage[T1]{fontenc}
  \usepackage[utf8]{inputenc}
  \usepackage{textcomp} 
\else 
  \usepackage{unicode-math} 
  \defaultfontfeatures{Scale=MatchLowercase}
  \defaultfontfeatures[\rmfamily]{Ligatures=TeX,Scale=1}
\fi
\usepackage{lmodern}
\ifPDFTeX\else
\fi
\IfFileExists{upquote.sty}{\usepackage{upquote}}{}
\IfFileExists{microtype.sty}{
  \usepackage[]{microtype}
  \UseMicrotypeSet[protrusion]{basicmath} 
}{}
\makeatletter
\@ifundefined{KOMAClassName}{
  \IfFileExists{parskip.sty}{%
    \usepackage{parskip}
  }{
    \setlength{\parindent}{0pt}
    \setlength{\parskip}{6pt plus 2pt minus 1pt}}
}{
  \KOMAoptions{parskip=half}}
\makeatother
\makeatletter
\ifx\paragraph\undefined\else
  \let\oldparagraph\paragraph
  \renewcommand{\paragraph}{
    \@ifstar
      \xxxParagraphStar
      \xxxParagraphNoStar
  }
  \newcommand{\xxxParagraphStar}[1]{\oldparagraph*{#1}\mbox{}}
  \newcommand{\xxxParagraphNoStar}[1]{\oldparagraph{#1}\mbox{}}
\fi
\ifx\subparagraph\undefined\else
  \let\oldsubparagraph\subparagraph
  \renewcommand{\subparagraph}{
    \@ifstar
      \xxxSubParagraphStar
      \xxxSubParagraphNoStar
  }
  \newcommand{\xxxSubParagraphStar}[1]{\oldsubparagraph*{#1}\mbox{}}
  \newcommand{\xxxSubParagraphNoStar}[1]{\oldsubparagraph{#1}\mbox{}}
\fi
\makeatother

\usepackage{longtable,booktabs,array}
\usepackage{calc} 
\usepackage{etoolbox}
\makeatletter
\patchcmd\longtable{\par}{\if@noskipsec\mbox{}\fi\par}{}{}
\makeatother
\IfFileExists{footnotehyper.sty}{\usepackage{footnotehyper}}{\usepackage{footnote}}
\makesavenoteenv{longtable}
\usepackage{graphicx}
\makeatletter
\newsavebox\pandoc@box
\newcommand*\pandocbounded[1]{
  \sbox\pandoc@box{#1}%
  \Gscale@div\@tempa{\textheight}{\dimexpr\ht\pandoc@box+\dp\pandoc@box\relax}%
  \Gscale@div\@tempb{\linewidth}{\wd\pandoc@box}%
  \ifdim\@tempb\p@<\@tempa\p@\let\@tempa\@tempb\fi
  \ifdim\@tempa\p@<\p@\scalebox{\@tempa}{\usebox\pandoc@box}%
  \else\usebox{\pandoc@box}%
  \fi%
}
\def\fps@figure{htbp}
\makeatother

\NewDocumentCommand\citeproctext{}{}
\NewDocumentCommand\citeproc{mm}{%
  \begingroup\def\citeproctext{#2}\cite{#1}\endgroup}
\makeatletter
 \let\@cite@ofmt\@firstofone
 \def\@biblabel#1{}
 \def\@cite#1#2{{#1\if@tempswa , #2\fi}}
\makeatother
\newlength{\cslhangindent}
\newlength{\csllabelwidth}
\newenvironment{CSLReferences}[2] 
 {\begin{list}{}{%
  \setlength{\itemindent}{0pt}
  \setlength{\leftmargin}{0pt}
  \setlength{\parsep}{0pt}
  \ifodd #1
   \setlength{\leftmargin}{\cslhangindent}
   \setlength{\itemindent}{-1\cslhangindent}
  \fi
  \setlength{\itemsep}{#2\baselineskip}}}
 {\end{list}}
\usepackage{calc}

\usepackage{fontspec}
\usepackage{microtype}  

\usepackage{booktabs}
\usepackage{longtable}
\usepackage{array}
\usepackage{tabularx}

\usepackage{graphicx}
\usepackage{float}

\usepackage{etoolbox}

\AtBeginEnvironment{longtable}{\small}

\usepackage[colorlinks=true, linkcolor=black, citecolor=black, urlcolor=blue!60!black]{hyperref}

\makeatletter
\def\ps@fancy{%
  \def\@oddhead{\small\textit{Which Leakage Types Matter?}\hfill\textit{Roth, 2026}}%
  \def\@evenhead{\small\textit{Which Leakage Types Matter?}\hfill\textit{Roth, 2026}}%
  \def\@oddfoot{\hfill\thepage\hfill}%
  \def\@evenfoot{\hfill\thepage\hfill}%
}
\makeatother

\newlength{\refhang}
\makeatletter
\@ifpackageloaded{caption}{}{\usepackage{caption}}
\AtBeginDocument{%
\ifdefined\contentsname
  \renewcommand*\contentsname{Table of contents}
\else
  \newcommand\contentsname{Table of contents}
\fi
\ifdefined\listfigurename
  \renewcommand*\listfigurename{List of Figures}
\else
  \newcommand\listfigurename{List of Figures}
\fi
\ifdefined\listtablename
  \renewcommand*\listtablename{List of Tables}
\else
  \newcommand\listtablename{List of Tables}
\fi
\ifdefined\figurename
  \renewcommand*\figurename{Figure}
\else
  \newcommand\figurename{Figure}
\fi
\ifdefined\tablename
  \renewcommand*\tablename{Table}
\else
  \newcommand\tablename{Table}
\fi
}
\@ifpackageloaded{float}{}{\usepackage{float}}
\floatstyle{ruled}
\@ifundefined{c@chapter}{\newfloat{codelisting}{h}{lop}}{\newfloat{codelisting}{h}{lop}[chapter]}
\floatname{codelisting}{Listing}

\makeatother
\makeatletter
\@ifpackageloaded{caption}{}{\usepackage{caption}}
\@ifpackageloaded{subcaption}{}{\usepackage{subcaption}}
\makeatother
\usepackage{bookmark}
\IfFileExists{xurl.sty}{\usepackage{xurl}}{} 
\hypersetup{
  pdftitle={Which Leakage Types Matter? },
  pdfauthor={Simon Roth},
  pdfkeywords={data leakage, cross-validation, reproducibility, effect
size, benchmark, meta-regression},
  colorlinks=true,
  linkcolor={blue},
  filecolor={Maroon},
  citecolor={Blue},
  urlcolor={Blue},
  pdfcreator={LaTeX via pandoc}}

\title{Which Leakage Types Matter?\\
\vspace{0.4em}\large\textnormal{A Quantitative Landscape Across 2,047 Benchmark Datasets}}
\author{Simon Roth}
\date{May 29, 2026}
\begin{document}
\maketitle
\begin{abstract}
Twenty-nine experiments (twenty-eight core, plus a boundary experiment
on 129 temporal datasets) across 2,047 iid tabular datasets support a
four-class taxonomy of data leakage organized by causal mechanism. Class
I (estimation leakage --- fitting scalers or encoders on full data) is
negligible: nine conditions all produce
\(|\Delta\text{AUC}| \leq 0.005\). Class II (selection leakage ---
peeking, seed cherry-picking, early stopping) is substantial at
practical dataset sizes (corpus median \(n \approx 1{,}900\)): the
measured effect is consistent with an approximately 90\%
noise-exploitation share (under the assumed zero-diversity decomposition
for seeds; see Limitation 11) inflating reported scores. Seed inflation
vanishes by \(n = 5{,}000\); peeking retains a residual at
\(n = 100{,}000\) that reflects genuine algorithm diversity, not
persistent leakage. Class III (memorization) is amplified by model
capacity: at 10\% duplication, six algorithms span \(d_z = 0.37\) (NB)
to \(1.11\) (DT), reaching \(d_z = 1.38\) for DT at 30\% duplication.
Class IV (boundary leakage) is invisible under random CV: a boundary
experiment across 129 temporal datasets (92 FOREX as null control, 14
with verified genuine timestamps, 23 with spurious time columns) shows
that random CV censors structural contamination. The pure temporal
effect averages +0.023 on genuine temporal datasets but near zero on
benchmarks without real drift. Feature selection leakage is negligible
at typical dimensionality but reaches +0.018 mean at high \(p/n\)
ratios, confirming Ambroise and McLachlan's (2002) finding at scale.
Metric selection flips model rankings on 31\% of datasets.
Cross-validation confidence intervals achieve only 55\% actual coverage
at nominal 95\% under the naive \(z\)-based default (fold SE divided by
\(\sqrt{k}\), \(z\) critical value), the convention practitioners report
by default; Conservative-Z (fold SD without \(\div\sqrt{k}\)) closes
most of the gap to approximately 87\%, but is not in standard practice.
Within the OpenML-class iid tabular regime studied here, the textbook
emphasis is inverted: the leakage type most prominent in standard
references (normalization) matters least; selection leakage at practical
dataset sizes matters most. Translation to non-tabular (image, text) and
other non-iid (group, spatial) domains is an open question (see
Limitations).
\end{abstract}

\newpage

\section{Introduction}\label{introduction}

Kapoor and Narayanan (\citeproc{ref-kapoor2023leakage}{2023}) audited
the machine learning literature across 17 scientific fields and found
294 published papers whose results were invalidated by data leakage
after publication; their living survey now catalogues 648 papers (as of
mid-2024) across 30 fields (\citeproc{ref-kapoor2025living}{Kapoor and
Narayanan 2025}). At benchmark scale, Recht et al.
(\citeproc{ref-recht2019imagenet}{2019}) documented the community-wide
pattern: a new ImageNet test set, constructed to match the original
collection protocol, exposed accuracy drops of 11--14 percentage points
across every model, evidence that years of repeated test-set reuse had
inflated the original measurements. The Kapoor and Recht findings are
complementary: per-paper leakage across the published literature, and
aggregate test-set adaptivity at the benchmark scale. Kapoor's taxonomy
identifies the types of leakage (preprocessing on full data, feature
leakage, overlap between training and test sets, temporal contamination)
but does not tell us which ones to worry about.

This distinction matters. A leakage type that inflates AUC by 0.001 on
average is a theoretical impurity. One that inflates it by +0.040 in
92\% of datasets is a practical crisis. The appropriate response (how
much engineering effort to invest in prevention, which leakage to audit
first, what to teach students) depends on magnitude, not merely on
existence. To my knowledge, no prior study has measured multiple leakage
types at scale on a shared corpus with a unified metric.

I fill this gap with twenty-nine experiments (twenty-eight core plus a
boundary experiment on temporal data), each a controlled perturbation of
a standard 5-fold cross-validation workflow, run across 2,047 binary
classification datasets from OpenML, PMLB, and ml. For every dataset and
every leakage type, I measure the AUC difference between the leaky
procedure and the clean procedure. The result is a quantitative
landscape of leakage effects. I embed internal validation (not
pre-registered, but built into the design): deterministic hashes split
the corpus at both the dataset level (discovery vs.~confirmation) and
the partition level, with content-addressed tracking throughout. All
testable effects replicate on the held-out confirmation split with zero
failures.

The central finding is that effect sizes span more than an order of
magnitude in raw \(\Delta\)AUC across leakage types (\(< 0.005\) for
estimation leakage to \(> 0.07\) for memorization leakage at extreme
settings), and the variation is predicted almost entirely by causal
mechanism. This is not a continuous spectrum. It is a categorical
distinction. Three classes emerge from the core experiments; a fourth
emerges from a boundary experiment on temporal data:

\textbf{Estimation leakage} (fitting a scaler, imputer, PCA, calibrator,
or outlier detector on the full dataset rather than on the training
fold) produces near-zero AUC inflation. Nine experiments (normalization,
PCA, calibration, chained preprocessing, outlier removal, feature
encoding, binning) all produce \(\Delta\)AUC \textless{} +0.005 (no
significant leakage inflation in any condition); the chained pipeline
(CE) reaches \(\Delta\)AUC = -0.0069, which is \emph{anti-leakage}:
per-fold preprocessing slightly outperforms full-data preprocessing
because each fold's pipeline adapts to that fold's distribution. The
bias is of order O(p/n) and vanishes at practical sample sizes.

\textbf{Selection leakage} (using holdout-set performance to guide model
selection, hyperparameter tuning, or seed choice across the split
boundary) produces \(\Delta\)AUC = +0.013 to +0.045 (d\(_z\) =
0.27--0.93). Four distinct mechanisms produce substantial effects at
practical dataset sizes: peeking at test-set performance to select from
\(k\) model configurations (\(\Delta\)AUC = +0.040, d\(_z\) = 0.93, 92\%
prevalence), cherry-picking the best random seed (\(\Delta\)AUC =
+0.045, d\(_z\) = 0.89, 92\% prevalence), using test data for early
stopping (\(\Delta\)AUC = +0.008, d\(_z\) = 0.46, 76\% prevalence), and
selecting from a screen of algorithms (\(\Delta\)AUC = +0.013, d\(_z\) =
0.269). Every selection mechanism decomposes into noise exploitation
(decaying as \(1/\sqrt{n}\)) and genuine diversity. Seed inflation
vanishes by \(n = 5{,}000\); peeking retains a diversity residual at
\(n = 100{,}000\).

\textbf{Memorization leakage} covers two mechanisms: (a) training on
exact or near-exact copies of evaluation rows (Exp H, cross-split
memorization), or (b) intra-split synthetic replicas of training
instances injected during class-imbalance correction (Exp G random
oversampling, Exp BA SMOTE). It produces the single largest raw effects
but depends on model capacity. Mechanism (a) crosses the split boundary;
mechanism (b) does not. Both depend on a model's capacity to fit
individual instances. Six algorithms span \(d_z =\) 0.37 (NB) to 1.11
(DT) at 10\% duplication, with a monotonic capacity ordering NB \(<\) LR
\(<\) XGB \(<\) RF \(<\) KNN \(<\) DT (at 10\% and 30\% duplication; at
5\%, KNN exceeds DT: DT's regularization-free memorization advantage
requires sufficient duplicate density to manifest).

\textbf{Boundary leakage}, where the partition strategy (random CV)
mismatches the deployment boundary (temporal, group, spatial), is
invisible under the standard protocol because random CV destroys the
structure that would reveal it. A boundary experiment across 129
temporal datasets shows the pure temporal effect averages +0.023 AUC on
datasets with genuine drift but near zero on benchmarks without real
temporal structure. The mechanism is distinct from Classes I--III: it is
not parameter averaging, not selection from K alternatives, not data
duplication. It is a structural mismatch between what the validation
procedure assumes about the data-generating process and what that
process actually is.

\subsection{Effect size metric and
notation}\label{effect-size-metric-and-notation}

Throughout this paper, I report two complementary metrics: the raw AUC
difference (\(\Delta\)AUC = AUC\(_{\text{leaky}}\) −
AUC\(_{\text{clean}}\)) and the standardized paired-difference effect
size d\(_z\). The latter is defined as
\(d_z = \bar{\Delta} / s_{\Delta}\), where \(s_{\Delta}\) is the
standard deviation of the within-subject differences
(\citeproc{ref-lakens2013calculating}{Lakens 2013}). The d\(_z\)
statistic can make small absolute effects sound large when
between-dataset variance is low, and d\(_z\) values are not directly
comparable across experiments run on different corpus subsets (the
denominator \(s_\Delta\) varies with corpus composition). I report both
scales so the reader can judge practical significance. For reference,
d\(_z\) = 0.93 at the corpus median corresponds to \(\Delta\)AUC
\(\approx\) 0.040, four hundredths of an AUC point. Whether this is
small or large depends on the decision context: in a Kaggle competition
where rank-ordering matters, +0.04 is decisive; in clinical screening or
fairness auditing where deployment depends on passing a fixed
performance threshold, +0.04 of phantom accuracy can mean deploying a
model that does not actually meet the bar.

\section{Related Work}\label{related-work}

\textbf{Leakage taxonomies.} Kapoor and Narayanan
(\citeproc{ref-kapoor2023leakage}{2023}) provide the most thorough
recent survey, identifying leakage in 294 papers across medicine,
biology, economics, and neuroscience; their living survey
(\citeproc{ref-kapoor2025living}{Kapoor and Narayanan 2025}) has since
grown to 648 papers across 30 fields. Their taxonomy classifies leakage
by the point in the pipeline where label information crosses the split
boundary: preprocessing, feature engineering, data overlap, and temporal
contamination. Kaufman et al. (\citeproc{ref-kaufman2012leakage}{2012})
cut it differently, by input-type legitimacy (illegitimate features vs
illegitimate training examples). The present taxonomy makes a third cut:
by the statistical mechanism that determines the magnitude (estimation,
selection, memorization). The three are complementary, answering
different questions: where in the pipeline, what input is illegitimate,
and which mechanism sets the magnitude.

\textbf{Individual leakage studies.} Several prior studies measure
specific leakage types in isolation. Ambroise and McLachlan
(\citeproc{ref-ambroise2002selection}{2002}) demonstrated that feature
selection on the full dataset before cross-validation produces severely
biased error estimates in gene expression studies. Guyon and Elisseeff
(\citeproc{ref-guyon2003introduction}{2003}) survey the broader feature
selection landscape, distinguishing filter, wrapper, and embedded
methods and reviewing variable-ranking and subset-selection strategies
across this special-issue collection. Varma and Simon
(\citeproc{ref-varma2006bias}{2006}) confirmed this for model selection,
showing that nested cross-validation is required for unbiased
estimation. Vandewiele et al.
(\citeproc{ref-vandewiele2021overly}{2021}) showed that SMOTE
(\citeproc{ref-chawla2002smote}{Chawla et al. 2002}) applied before
splitting inflates AUC on imbalanced datasets. Kaufman et al.
(\citeproc{ref-kaufman2012leakage}{2012}) formalized data leakage as a
target-information legitimacy problem, with clinical prediction among
their canonical examples. Sasse et al.
(\citeproc{ref-sasse2025featuretarget}{2025}) provide a comprehensive
overview of leakage scenarios in supervised learning, including
feature-to-target leakage as a distinct class. Cawley and Talbot
(\citeproc{ref-cawley2010overfitting}{2010}) analysed the bias from
hyperparameter selection on the validation set and showed that selecting
over a configuration grid systematically over-fits the model-selection
criterion; Bergstra and Bengio (\citeproc{ref-bergstra2012random}{2012})
showed that random search is more efficient than grid search; the
selection-bias-from-reporting-the-best mechanism applies regardless of
search strategy; Bischl et al. (\citeproc{ref-bischl2023hpo}{2023})
provide a comprehensive survey of hyperparameter optimization, including
the interaction between tuning strategy and evaluation bias. Hastie,
Tibshirani, and Friedman (\citeproc{ref-hastie2009elements}{2009}, Ch.
7) provide the theoretical framework for model assessment versus model
selection, from which the estimation leakage bound O(p/n) follows.

One recent study comes closest to a multi-type comparison. Rosenblatt et
al. (\citeproc{ref-rosenblatt2024data}{2024}) measure five leakage forms
on shared neuroimaging datasets with a unified metric and find that
leakage severity depends on dataset size, but their mechanisms are
domain-specific (site correction, family structure) and the corpus is
connectome data, not general-purpose tabular data. This study's
high-dimensional wrapper-feature-selection result (+0.018 mean
\(\Delta\)AUC at \(p/n >\) 0.1, \(N\) = 49 datasets) aligns with
Ambroise and McLachlan (\citeproc{ref-ambroise2002selection}{2002})'s
foundational gene-expression finding. To my knowledge, no prior work
measures all four mechanism classes in a single controlled experiment on
a shared corpus with a unified metric.

\textbf{Practitioner guides.} Raschka
(\citeproc{ref-raschka2020modelevaluation}{2020}) provides a thorough
tutorial on model evaluation and selection methodology, covering the
statistical foundations of cross-validation, bootstrapping, and nested
resampling. Lones (\citeproc{ref-lones2024pitfalls}{2024}) catalogues
common ML pitfalls including several leakage-adjacent issues. Both works
provide qualitative guidance on leakage avoidance; this study
complements them by quantifying which pitfalls matter most.

\textbf{Detection and prevention tools.} Since 2024, automated leakage
detection has accelerated. Urban, Subotić, and Drobnjaković
(\citeproc{ref-drobnjakovic2025abstract}{2025}) applied abstract
interpretation to ML notebook code, tracking partition-membership labels
through data-flow operations at 93\% precision. Truong et al.
(\citeproc{ref-truong2025leakagedetector2}{2025}) extended
LeakageDetector to Jupyter pipelines with LLM-driven corrections.
Apicella, Isgrò, and Prevete (\citeproc{ref-apicella2025button}{2025})
documented how black-box ML tool usage amplifies leakage risk in
transfer learning. These tools detect leakage after the fact; the
present study quantifies how much each type matters.

\textbf{Adaptive data analysis.} Dwork et al.
(\citeproc{ref-dwork2015reusable}{2015}) formalized the dangers of
reusing holdout data for multiple adaptive decisions, showing that
validity degrades with repeated access. Their ``reusable holdout''
framework (which uses differential-privacy mechanisms to allow
controlled repeated access with quantified accuracy guarantees) provides
a theoretical foundation for why peeking produces O(1) bias: each
adaptive decision using the same evaluation set transfers information
from that set into the model, regardless of sample size. The assess-once
constraint proposed in the companion grammar
(\citeproc{ref-roth2026grammar}{Roth 2026}) is the zero-reuse limit of
this framework (\(\epsilon = 0\)): rather than permitting controlled
repeated access, it permits no reuse within a session, trading the
reusable holdout's flexibility for enforceability in a static type
system.

\textbf{Cross-validation uncertainty.} Arlot and Celisse
(\citeproc{ref-arlot2010survey}{2010}) provide the definitive survey of
cross-validation procedures. Bengio and Grandvalet
(\citeproc{ref-bengio2004noUnbiased}{2004}) proved that no
\emph{universal} (distribution-free) unbiased estimator of
cross-validation variance exists, and Nadeau and Bengio
(\citeproc{ref-nadeau2003inference}{2003}) proposed a corrected variance
estimate accounting for the non-independence of overlapping folds.
Dietterich (\citeproc{ref-dietterich1998approximate}{1998}) compared
five statistical tests for algorithm comparison, establishing that
paired tests on shared folds are anti-conservative. Bates, Hastie, and
Tibshirani (\citeproc{ref-bates2024crossvalidation}{2024}) showed that
standard k-fold confidence intervals can have coverage far below the
nominal level, with simulated miscoverage rates two to three times the
desired rate. Varoquaux
(\citeproc{ref-varoquaux2018crossvalidation}{2018}) demonstrated
\(\pm 10\%\) error bars at n = 100 in neuroimaging studies. Tsamardinos,
Greasidou, and Borboudakis
(\citeproc{ref-tsamardinos2018bootstrap}{2018}) proposed bootstrapping
out-of-sample predictions as an alternative with improved coverage
properties. I measure this coverage gap empirically across 2,047
datasets (Experiment AO).

\textbf{Dataset moderators.} Rosenblatt et al.
(\citeproc{ref-rosenblatt2024data}{2024}) showed that small datasets
amplify leakage severity in neuroimaging. My Bayesian meta-regression
(Section~\ref{sec-meta-regression}) extends this analysis to 2,047
general-purpose datasets.

\textbf{Benchmark studies.} OpenML
(\citeproc{ref-vanschoren2013openml}{Vanschoren et al. 2013};
\citeproc{ref-bischl2025openml}{Bischl et al. 2025}) provides a
standardized repository of classification datasets used in
reproducibility research. I draw 2,047 datasets from OpenML, PMLB, and
ml, covering binary problems with 100 to 946,799 rows.

\section{Experimental Design}\label{experimental-design}

\subsection{Dataset corpus}\label{dataset-corpus}

I collect all binary classification datasets from OpenML with at least
100 rows and at least 2 features, excluding datasets flagged as inactive
or duplicate (2,053 datasets), supplemented by 119 curated datasets from
PMLB (Penn Machine Learning Benchmarks) and 116 from the ml package.
After deduplication by content hash, the corpus contains 2,288 unique
datasets, of which 2,047 completed automatic processing successfully
(89.5\%). The remaining 241 were excluded due to loading errors (118) or
filtering criteria (123: too few rows, single class, or excessive
dimensionality).

The corpus spans four orders of magnitude in sample size (median \(n =\)
1,901, max = 946,799), two orders of magnitude in feature count (median
\(p =\) 18), and diverse data characteristics. The corpus includes 95
datasets with \textgreater100K rows; stratified analysis
(Section~\ref{sec-capacity}) confirms these do not distort key findings.

\subsection{Internal validation
protocol}\label{internal-validation-protocol}

Before running experiments, I assign each unique dataset to a
\emph{discovery} split or a \emph{confirmation} split via a
deterministic hash of the dataset name and source (50/50 allocation,
\(\approx 1{,}020\) per split). Hypotheses formulated on discovery-split
results are tested on confirmation-split results. \textbf{This is an
internal validation protocol, not a formal pre-registration.} I did not
register hypotheses on an external platform (e.g., OSF). I characterize
the study as exploratory with built-in internal validation and
directional prediction tracking. Pre-registration is uncommon in ML
methodology papers, but the corpus, baseline, and effect-size
operationalization here are all author-curated: without recording
directional predictions before observing results, post-hoc
rationalization would be indistinguishable from confirmation, and the
taxonomy's empirical claims would not be independently falsifiable.

For 13 of the 28 experiments, I recorded directional predictions
(expected class membership and effect size range) before data
collection. The prediction scorecard (Section~\ref{sec-scorecard})
reports 10 of 13 confirmed.

\subsection{Leakage experiments}\label{leakage-experiments}

The design is a within-subject counterfactual experiment: each dataset
and model serves as its own control, and both the clean and leaky
outcomes are observed under identical fold assignments, eliminating
between-dataset variance from the treatment estimate. A \emph{clean}
workflow (methodologically correct) and a \emph{leaky} workflow
(containing a single, controlled perturbation) are run on the same data
with the same 5-fold stratified cross-validation and the same random
seed. The only difference is the leakage perturbation. The effect is the
paired AUC difference: \(\Delta\)AUC = AUC\(_{\text{leaky}}\) −
AUC\(_{\text{clean}}\). Because the treatment is a code-path switch
(fully reversible at the unit level) and the unit (the dataset) is
identical across runs, both potential outcomes are observable for every
(dataset, model) pair; the fundamental problem of causal inference does
not apply to computational experiments. I report the corpus distribution
of paired differences across all (dataset, model) pairs rather than
collapsing to a single corpus-mean, because the heterogeneity across
datasets is the substantive finding. Several experiments extend this to
factorial designs (e.g., algorithm \(\times\) duplication rate in
Experiment H) or dose-response curves (e.g., number of seeds
\(K = 5\)--\(100\) in Experiment AP, subsample size
\(n = 50\)--\(10{,}000\) in Experiment AN). The primary algorithms are
logistic regression (LR) and random forest (RF) from scikit-learn
(\citeproc{ref-pedregosa2011scikit}{Pedregosa et al. 2011}), chosen for
their ubiquity and contrasting capacity. Four additional algorithms (NB,
XGB, KNN, DT) are tested where algorithm capacity is the variable of
interest.

The clean baseline is a correct 5-fold cross-validation workflow, not an
oracle (the true generalization performance under the population
data-generating distribution: computable on synthetic data, unobservable
on real data). The clean baseline differs from the oracle in two
directions. First, every 5-fold CV workflow that reports the best
CV-mean across multiple candidate models has performed implicit
selection on the validation folds, inflating the baseline's own
performance estimate (\citeproc{ref-cawley2010overfitting}{Cawley and
Talbot 2010}); the measured \(\Delta\)AUC = leaky − clean therefore
captures the \emph{additional} inflation from the leakage perturbation
on top of the baseline's residual selection bias. Second, on datasets
with structural correlation (groups, temporal, spatial), random folding
scatters correlated observations across folds, and the ``clean''
baseline already benefits from this structural leakage; the boundary
experiment on 129 temporal datasets (Section~\ref{sec-boundary}) shows
this affects the iid majority near-zero but substantially where real
distribution drift exists. The reported Class II \(\Delta\)AUC values
are conservative lower bounds relative to an oracle evaluation. All AUC
values are computed from out-of-fold predictions: each observation is
scored only by a model that never saw it during training, so all
downstream statistics operate on held-out performance without
circularity.

The twenty-eight core experiments plus a boundary experiment span five
categories:

\textbf{Estimation (11 experiments).} A (normalization), A2
(multi-algorithm normalization), A3 (scaler comparison), D (test
contamination), E (outlier removal), F (feature encoding), T (binning),
Q (high-cardinality vocabulary: CountVectorizer fit on full data vs
per-fold), CE (chained estimation), AB (PCA), AF (calibration).

\textbf{Selection (8 experiments).} B (peeking at k = 1--19), K
(algorithm baseline), BB (early stopping), AI (seed inflation,
best-of-10), AQ (screen inflation at K = 1--11), AC (target encoding,
categorical datasets only), AN (n-scaling across eight subsample sizes,
N = 493), AP (seed stability dose-response at K = 5--100, N = 1,965).

\textbf{Memorization (3 experiments).} G (oversampling), H (duplicates
at 5--30\%), BA (SMOTE vs random oversampling).

\textbf{Boundary (2 experiments).} P (grouped splits: random CV vs
GroupKFold on a click-prediction dataset with \texttt{ad\_id} as group
column), and the temporal boundary experiment across 129 datasets (92
FOREX null control, 14 genuine temporal, 23 spurious).

\textbf{Null and diagnostic (5 experiments).} J (compound), L (CV
variance), AK (stack meta-leakage), AE (seed noise floor), AO (CV
coverage gap: nominal vs actual CI coverage, N = 2,047).

\subsection{Effect size metric}\label{sec-effect-size}

For each experiment and dataset, I compute the AUC difference:
\(\Delta\)AUC = AUC\(_{\text{leaky}}\) − AUC\(_{\text{clean}}\). Across
datasets, I report d\(_z\) = \(\bar{\Delta}\) / \(s_{\Delta}\) as the
standardized paired-difference effect size
(\citeproc{ref-lakens2013calculating}{Lakens 2013}), with 95\%
confidence intervals from a \(t\)-distribution with \(N - 1\) degrees of
freedom (where \(N\) is the number of datasets in the experiment, not
the number of CV folds). I also report raw \(\Delta\)AUC with confidence
intervals for interpretability. These between-dataset paired-difference
CIs are statistically distinct from the per-dataset CV-fold CIs analyzed
in Section~\ref{sec-cv-coverage}; the CV-coverage gap (Experiment AO)
does not apply to the \(d_z\) and \(\Delta\)AUC intervals reported
throughout this paper. Secondary metrics: proportion of datasets with
\(\Delta\)AUC \textgreater{} 0 (prevalence of positive inflation), and
median \(\Delta\)AUC (robustness to outliers). The detection floor at N
= 2,047 is d\(_z\) = 0.057. With 29 experiments, multiple testing is a
concern. Applying Benjamini-Hochberg false discovery rate (FDR)
correction (\citeproc{ref-benjamini1995controlling}{Benjamini and
Hochberg 1995}) at \(\alpha = 0.05\) does not change any conclusion:
Class II and III effects have d\(_z\) \textgreater{} 0.3 (all
\(p < 10^{-8}\)), surviving any reasonable correction threshold. The
Class I null claims rest on effect sizes below the noise floor, not on
\(p\)-values.

\subsection{Measurement noise floor}\label{measurement-noise-floor}

Experiment AE serves as a placebo control: both pipelines are
methodologically clean with different random seeds, so any observed
difference is pure measurement noise. Logistic regression is perfectly
deterministic (cross-seed SD = 0.000). Random forests show median
cross-seed SD = 0.0027, with a 10.4\% winner-flip rate (the probability
that a different random seed reverses which algorithm appears better).
Any effect with \textbar{}\(\Delta\)AUC\textbar{} \textless{} 0.003 or
\textbar d\(_z\)\textbar{} \textless{} 0.06 could be a seed artifact.
All results below are interpreted against this floor.

\section{Results}\label{results}

\begin{figure}

\centering{

\includegraphics[width=1\linewidth,height=\textheight,keepaspectratio]{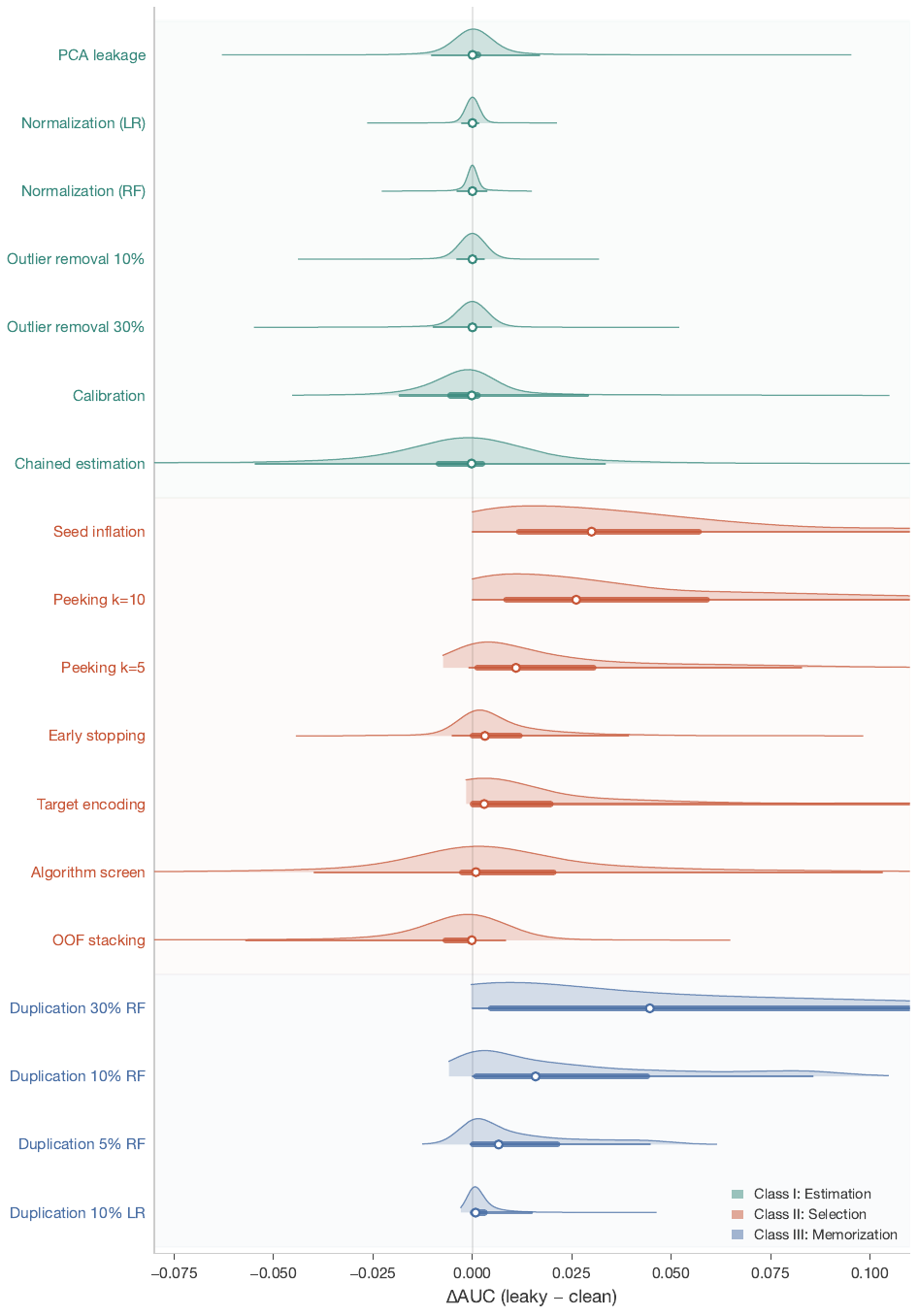}

}

\caption{\label{fig-raincloud}Distribution of \(\Delta\)AUC across
leakage experiments, grouped by leakage class. Class I (estimation,
teal) centers on zero. Class II (selection, red) shows persistent
positive inflation. Class III (memorization, blue) leakage varies with
model capacity and duplication rate.}

\end{figure}%

\subsection{Estimation leakage produces near-zero
effects}\label{estimation-leakage-produces-near-zero-effects}

Eleven experiments test estimation leakage across different
preprocessing operations. Nine representative conditions are reported
below; all produce near-zero raw effects.

\begin{longtable}[]{@{}
  >{\raggedright\arraybackslash}p{(\linewidth - 8\tabcolsep) * \real{0.4247}}
  >{\centering\arraybackslash}p{(\linewidth - 8\tabcolsep) * \real{0.1096}}
  >{\centering\arraybackslash}p{(\linewidth - 8\tabcolsep) * \real{0.1096}}
  >{\centering\arraybackslash}p{(\linewidth - 8\tabcolsep) * \real{0.1096}}
  >{\raggedright\arraybackslash}p{(\linewidth - 8\tabcolsep) * \real{0.2466}}@{}}
\caption{Class I estimation leakage
results.}\label{tbl-class1}\tabularnewline
\toprule\noalign{}
\begin{minipage}[b]{\linewidth}\raggedright
Experiment
\end{minipage} & \begin{minipage}[b]{\linewidth}\centering
d\(_z\)
\end{minipage} & \begin{minipage}[b]{\linewidth}\centering
\(\Delta\)AUC
\end{minipage} & \begin{minipage}[b]{\linewidth}\centering
N
\end{minipage} & \begin{minipage}[b]{\linewidth}\raggedright
Note
\end{minipage} \\
\midrule\noalign{}
\endfirsthead
\toprule\noalign{}
\begin{minipage}[b]{\linewidth}\raggedright
Experiment
\end{minipage} & \begin{minipage}[b]{\linewidth}\centering
d\(_z\)
\end{minipage} & \begin{minipage}[b]{\linewidth}\centering
\(\Delta\)AUC
\end{minipage} & \begin{minipage}[b]{\linewidth}\centering
N
\end{minipage} & \begin{minipage}[b]{\linewidth}\raggedright
Note
\end{minipage} \\
\midrule\noalign{}
\endhead
\bottomrule\noalign{}
\endlastfoot
A: Normalization, LR & -0.02 & 0.000 & 2,047 & Below noise floor \\
A: Normalization, RF & -0.05 & 0.000 & 2,047 & \\
E: Outlier removal 10\% & -0.03 & 0.000 & 2,047 & \\
E: Outlier removal 30\% & 0.00 & 0.000 & 2,047 & \\
F: Feature encoding, LR & +0.01 & +0.0006 & 1,029 & Categorical only \\
T: Binning, LR & +0.05 & +0.0004 & 1,063 & \\
CE: Chained pipeline & -0.15 & -0.007 & 2,047 & Per-fold slightly
\emph{better} \\
AB: PCA & 0.08 & 0.001 & 1,930 & Below noise floor \\
AF: Calibration & 0.05 & 0.001 & 2,047 & Below noise floor \\
\end{longtable}

On iid tabular benchmarks at typical dataset sizes, fitting a scaler on
the full dataset before splitting produces negligible bias due to
leakage. The bias is of order O(p/n), which at the corpus median
(\(p =\) 18, \(n =\) 1,901) produces \(\sim 0.001\) AUC per feature,
below detection threshold.

The chained estimation experiment (CE) deserves special note. I tested a
complete pipeline (global scaling + global encoding + global PCA)
stacked as a chain of estimation leakages. The result is d\(_z\) = -0.15
(\(\Delta\)AUC = -0.007): per-fold preprocessing produces slightly
\emph{better} estimates than the leaked chain. The direction is
counterintuitive but the interpretation is straightforward: estimation
on more data (train + test) produces a marginally better scaler, but the
clean procedure benefits from adapting the pipeline to each fold's
specific data distribution.

\subsection{Selection leakage is the dominant
effect}\label{selection-leakage-is-the-dominant-effect}

Four distinct selection mechanisms produce substantial effects at
practical dataset sizes.

\subsubsection{Peeking at test labels (Exp
B)}\label{peeking-at-test-labels-exp-b}

Peeking leakage (selecting the best of \(k\) model configurations by
their test-set performance rather than by honest cross-validation) is
the most universal effect in the landscape. The pool contains 19
configurations (9 random forests, 5 logistic regressions, 5 decision
trees with varying hyperparameters); the leaky workflow picks the best
of a random \(k\)-subset evaluated on the test set, while the clean
baseline picks one configuration at random.

\begin{longtable}[]{@{}
  >{\centering\arraybackslash}p{(\linewidth - 6\tabcolsep) * \real{0.2500}}
  >{\centering\arraybackslash}p{(\linewidth - 6\tabcolsep) * \real{0.2500}}
  >{\centering\arraybackslash}p{(\linewidth - 6\tabcolsep) * \real{0.2500}}
  >{\centering\arraybackslash}p{(\linewidth - 6\tabcolsep) * \real{0.2500}}@{}}
\toprule\noalign{}
\begin{minipage}[b]{\linewidth}\centering
Configurations evaluated (\(k\))
\end{minipage} & \begin{minipage}[b]{\linewidth}\centering
d\(_z\)
\end{minipage} & \begin{minipage}[b]{\linewidth}\centering
\(\Delta\)AUC {[}95\% CI{]}
\end{minipage} & \begin{minipage}[b]{\linewidth}\centering
\(P(\Delta > 0)\)
\end{minipage} \\
\midrule\noalign{}
\endhead
\bottomrule\noalign{}
\endlastfoot
1 & -0.56 & -0.021 & 0.20 \\
2 & -0.08 & -0.002 & 0.33 \\
5 & +0.71 & +0.022 & 0.84 \\
10 & +0.93 & +0.040 & 0.92 \\
15 & +0.93 & +0.043 & 0.92 \\
19 & +0.94 & +0.044 & 0.92 \\
\end{longtable}

At \(k = 10\), the mean AUC inflation is +0.040 (d\(_z\) = 0.93 {[}0.88,
0.98{]}), with positive inflation in 92\% of datasets.

\textbf{A non-monotonicity at \(k = 1\).} The existing literature on
model selection bias treats inflation as uniformly positive (Ambroise
and McLachlan (\citeproc{ref-ambroise2002selection}{2002}); Varma and
Simon (\citeproc{ref-varma2006bias}{2006})); I find a signed reversal.
At \(k = 1\), peeking \emph{decreases} performance (d\(_z\) = -0.56,
inflation in only 20\% of datasets); at \(k = 2\), the effect is near
zero (d\(_z\) = -0.08). This is not a mechanistic harm: at \(k=1\), the
leaky-vs-clean contrast is a zero-mean comparison by construction (both
pick one configuration), and the leaky side reduces to a single noisy
test-set AUC against the clean side's stabler CV baseline, so the
negative d\(_z\) is single-shot noise relative to a more stable
comparator, or sampling fluctuation around zero. The practitioner-facing
implication is the same: a researcher evaluating only a few
configurations may observe no inflation and conclude the procedure is
safe. At \(k \geq 5\), the order-statistic effect (maximum of \(k\)
draws) dominates and inflation becomes systematic.

\begin{figure}

\centering{

\includegraphics[width=0.8\linewidth,height=\textheight,keepaspectratio]{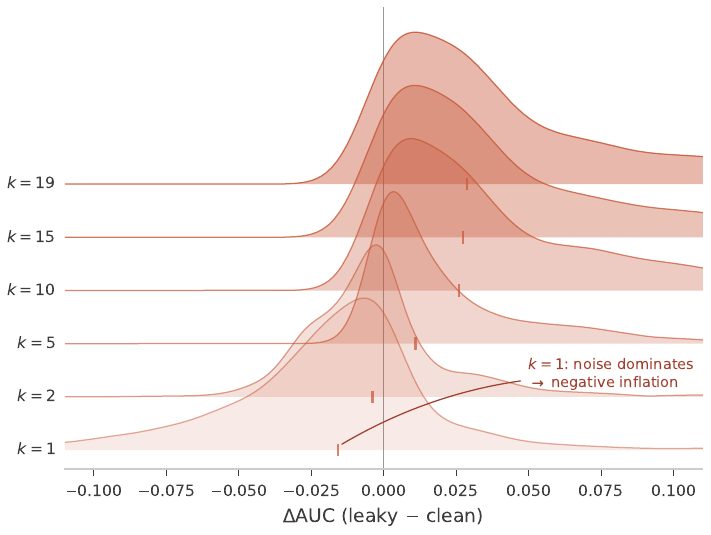}

}

\caption{\label{fig-peeking}Peeking inflation distribution across 2,047
datasets at \(k = 10\). The distribution is right-skewed with 92\%
positive prevalence.}

\end{figure}%

\textbf{Sample size correlation.} The correlation between peeking
inflation and \(\log(n)\) is \(r =\) -0.04: near zero. Peeking does not
diminish on larger datasets. Stratifying by dataset size confirms this;
the peeking effect size remains large and stable across all strata:
small datasets (\textless500 rows) d\(_z\) = 0.91, medium (500--5K)
d\(_z\) = 0.94, large (5K+) d\(_z\) = 0.94. The finding is confirmed on
the held-out split: discovery d\(_z\) = 0.94, confirmation d\(_z\) =
0.92.

\subsubsection{Seed cherry-picking (Exp AI,
AP)}\label{seed-cherry-picking-exp-ai-ap}

Seed inflation (reporting the best AUC across multiple random seeds
rather than the mean) produces d\(_z\) = +0.89 (92\% of datasets
affected). The mean AUC inflation from best-of-10 seeds is +0.045,
comparable in magnitude to peeking.

Experiment AP extends this to a dose-response design across K = 5 to 100
seeds (N = 1,965 unique datasets):

\begin{longtable}[]{@{}ccc@{}}
\toprule\noalign{}
K seeds & LR \(\Delta\)AUC & RF \(\Delta\)AUC \\
\midrule\noalign{}
\endhead
\bottomrule\noalign{}
\endlastfoot
5 & 0.000 & 0.012 \\
10 & 0.000 & 0.016 \\
25 & 0.000 & 0.021 \\
50 & 0.000 & 0.024 \\
100 & 0.000 & 0.026 \\
\end{longtable}

\newpage

\begin{figure}

\centering{

\includegraphics[width=0.7\linewidth,height=\textheight,keepaspectratio]{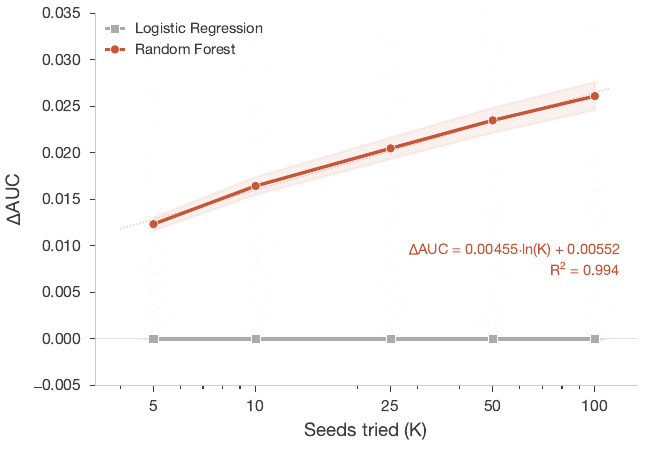}

}

\caption{\label{fig-seed}Seed inflation dose-response: RF inflation
grows logarithmically with K seeds while LR remains deterministic at
zero.}

\end{figure}%

Two observations. First, logistic regression is perfectly deterministic
(inflation = 0.000 at all K), as expected from its convex loss surface.
The seed effect is entirely an artifact of stochastic selection.

Second, RF seed inflation follows a logarithmic dose-response:

\[\Delta\text{AUC}(K) = 0.00455 \cdot \log(K) + 0.00552 \quad (R^2 = 0.994)\]

The regression is fit to 5 dose levels (\(K = 5, 10, 25, 50, 100\)),
each aggregated over 1,965 datasets. \(R^2\) is computed on the 5
aggregated means, not on 1,965 individual observations. The logarithmic
form is an empirical fit (\(R^2 > 0.99\)), not a derived law.
Extrapolation beyond the tested range (\(K = 5\)--\(100\)) is
speculative, but the logarithmic trend means the inflation only grows
worse. A researcher reporting ``best of 100 seeds'' is not engaging in
harmless hyperparameter tuning; the bias is predictable and monotonic in
\(K\).

\subsubsection{Early stopping on test data (Exp
BB)}\label{early-stopping-on-test-data-exp-bb}

Early stopping (using the test set as the convergence criterion for
iterative algorithms) produces d\(_z\) = +0.46 (\(\Delta\)AUC = +0.008,
N = 2,047), with positive inflation in 76\% of datasets. Each iteration
that observes the test-set loss transfers label information into the
model's stopping point.

\subsubsection{Screen selection (Exp AQ)}\label{screen-selection-exp-aq}

Algorithm screening (selecting the best-performing algorithm from a
pool) produces d\(_z\) = +0.27 (\(\Delta\)AUC = +0.013, N = 2,047). The
screen inflation is K-invariant:

\begin{longtable}[]{@{}ccc@{}}
\toprule\noalign{}
K algorithms & Mean \(\Delta\)AUC & d\(_z\) \\
\midrule\noalign{}
\endhead
\bottomrule\noalign{}
\endlastfoot
1 & +0.013 & +0.27 \\
5 & +0.013 & +0.27 \\
11 & +0.013 & +0.27 \\
\end{longtable}

The numerical identity across K = 1, 5, 11 reflects mechanism, not
measurement coincidence. Two parts: (i) the K=1 row is non-zero because
a \emph{single} screen evaluation against the test holdout is already
selection bias (relative to the random-of-1 clean baseline); and (ii)
additional candidates from a correlated pool produce no empirically
detectable growth, because all K algorithms evaluated on the same CV
folds fit the same signal in the same data; their errors are highly
correlated, so the effective number of independent draws stays near 1
regardless of \(K\) and the maximum barely grows. The bias is
empirically constant across K = 1 to 11; extrapolation to K \(\gg\) 11
would predict mild additional growth at the correlation-attenuated rate,
but the experiment does not test that range.

Seed inflation (Exp AP) shows the opposite pattern. For stochastic
algorithms (RF in my experiments), different seeds produce sufficient
prediction variance that the effective number of independent draws grows
with \(K\), so the maximum grows logarithmically. For deterministic
algorithms (LR), seed inflation is zero (the convex optimizer produces
the same model regardless of seed). Whether the variance reflects
genuinely independent models or merely different boundary-case
alignments is outside this paper's scope; the empirical pattern is
logarithmic growth for stochastic algorithms.

The K-invariance does not mean screening is harmless: d\(_z\) = +0.27 is
real bias from a single reporting decision, and in practice screening
compounds with tuning, feature selection, and other pipeline choices
that each contribute their own selection pressure. The bias comes from
the single act of evaluating on held-out data and picking the winner,
not from the number of candidates.

\subsubsection{Target encoding (Exp AC)}\label{target-encoding-exp-ac}

Target encoding (computing the target-class mean per category on the
full dataset, including test rows) produces d\(_z\) = +0.46 on the 1,208
datasets with categorical features (80\% prevalence, \(\Delta\)AUC =
+0.021).

This is mechanistically distinct from ordinal feature encoding (Exp F:
d\(_z\) = +0.01), which assigns integer codes without using label
information. Target encoding uses label information to construct the
encoding: each category value becomes its conditional target probability
estimated from the full data. The classification of this experiment is a
hybrid case: by \emph{causal mechanism} (fitting a statistic on full
data and leaking encoder state into evaluation), the pathway is Class I
(estimation leakage with a label-conditioned statistic); by
\emph{behavioral magnitude}, the inflation matches Class II (the
information content of a label-conditional statistic is much larger than
a marginal feature statistic). I assign it to Class II for behavioral
consistency, but flag this as a partial exception to the mechanism-first
rule: the correct grammar-level remedy is Constraint 2 (fit
transformations inside fold, not Constraint 1 (assess-once). A reader
who would prefer a strict mechanism-only taxonomy should reclassify this
experiment as a Class I-L (label-conditioned estimation) sub-type; the
empirical results are unchanged either way.

\subsection{Memorization leakage is amplified by memorization
capacity}\label{memorization-leakage-is-amplified-by-memorization-capacity}

Duplicate leakage (Exp H) produces a clear dose-response with a striking
algorithm gap:

\begin{longtable}[]{@{}
  >{\centering\arraybackslash}p{(\linewidth - 12\tabcolsep) * \real{0.1429}}
  >{\centering\arraybackslash}p{(\linewidth - 12\tabcolsep) * \real{0.1429}}
  >{\centering\arraybackslash}p{(\linewidth - 12\tabcolsep) * \real{0.1429}}
  >{\centering\arraybackslash}p{(\linewidth - 12\tabcolsep) * \real{0.1429}}
  >{\centering\arraybackslash}p{(\linewidth - 12\tabcolsep) * \real{0.1429}}
  >{\centering\arraybackslash}p{(\linewidth - 12\tabcolsep) * \real{0.1429}}
  >{\centering\arraybackslash}p{(\linewidth - 12\tabcolsep) * \real{0.1429}}@{}}
\caption{Memorization leakage dose-response across six algorithms,
ordered by model capacity (\(N\) = 2,047 for LR/RF; \(N\) = 2,005 for
DT/KNN; \(N\) = 989 for XGB/NB). The capacity ordering NB \(<\) LR \(<\)
XGB \(<\) RF \(<\) KNN \(<\) DT is monotonic.}\tabularnewline
\toprule\noalign{}
\begin{minipage}[b]{\linewidth}\centering
Duplication rate
\end{minipage} & \begin{minipage}[b]{\linewidth}\centering
NB d\(_z\)
\end{minipage} & \begin{minipage}[b]{\linewidth}\centering
LR d\(_z\)
\end{minipage} & \begin{minipage}[b]{\linewidth}\centering
XGB d\(_z\)
\end{minipage} & \begin{minipage}[b]{\linewidth}\centering
RF d\(_z\)
\end{minipage} & \begin{minipage}[b]{\linewidth}\centering
KNN d\(_z\)
\end{minipage} & \begin{minipage}[b]{\linewidth}\centering
DT d\(_z\)
\end{minipage} \\
\midrule\noalign{}
\endfirsthead
\toprule\noalign{}
\begin{minipage}[b]{\linewidth}\centering
Duplication rate
\end{minipage} & \begin{minipage}[b]{\linewidth}\centering
NB d\(_z\)
\end{minipage} & \begin{minipage}[b]{\linewidth}\centering
LR d\(_z\)
\end{minipage} & \begin{minipage}[b]{\linewidth}\centering
XGB d\(_z\)
\end{minipage} & \begin{minipage}[b]{\linewidth}\centering
RF d\(_z\)
\end{minipage} & \begin{minipage}[b]{\linewidth}\centering
KNN d\(_z\)
\end{minipage} & \begin{minipage}[b]{\linewidth}\centering
DT d\(_z\)
\end{minipage} \\
\midrule\noalign{}
\endhead
\bottomrule\noalign{}
\endlastfoot
5\% & +0.29 & +0.34 & +0.61 & +0.81 & +0.92 & +0.87 \\
10\% & +0.37 & +0.44 & +0.78 & +0.90 & +1.01 & +1.11 \\
30\% & +0.42 & +0.48 & +0.86 & +0.95 & +1.25 & +1.38 \\
\end{longtable}

\begin{figure}

\centering{

\includegraphics[width=0.6\linewidth,height=\textheight,keepaspectratio]{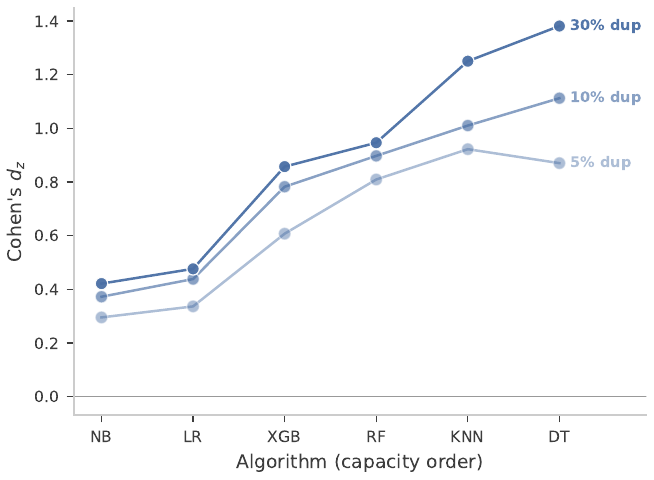}

}

\caption{\label{fig-capacity}Capacity amplification. Each line connects
six algorithms ordered by capacity (NB → LR → XGB → RF → KNN → DT) at a
fixed duplication rate. Higher duplication shifts all algorithms upward
(intercept), but the lines also fan out: the gap between constrained
(LR) and flexible (DT) models widens from \(\Delta\)AUC = 0.011 at 5\%
to 0.064 at 30\%, revealing a capacity \(\times\) duplication
interaction.}

\end{figure}%

The capacity ordering NB \(<\) LR \(<\) XGB \(<\) RF \(<\) KNN \(<\) DT
is consistent with a general amplification principle: memorization
leakage scales with a model's ability to overfit individual training
instances. Gaussian Naive Bayes, constrained by its class-conditional
independence assumption, shows the smallest effects (d\(_z\) = 0.37 at
10\%)---even below logistic regression (d\(_z\) = 0.44). XGBoost
(d\(_z\) = 0.78), despite its high predictive capacity, is regularized
by default (max\_depth = 6, L2 penalty), limiting memorization. Random
forests (d\(_z\) = 0.90) reduce per-tree memorization through bagging.
K-nearest neighbors (d\(_z\) = 1.01) memorize through instance storage:
duplicated test rows become their own nearest neighbors. Unconstrained
decision trees (d\(_z\) = 1.11) achieve the highest memorization by
fitting every training instance exactly. The full Class III range across
six algorithms is d\(_z\) = 0.29 (NB, 5\%) to 1.38 (DT, 30\%).

\textbf{SMOTE descriptively matches random oversampling (Exp BA).} On
777 datasets with class imbalance, SMOTE oversampling before splitting
produces d\(_z\) = +0.55, and random oversampling produces d\(_z\) =
+0.56. The two methods produce descriptively indistinguishable leakage
(mean, SD, IQR, skew, and kurtosis all match). SMOTE's synthetic
interpolation offers no descriptive protection against memorization
leakage in this corpus.

\subsection{Compound effects are
sub-additive}\label{compound-effects-are-sub-additive}

Experiment J stacks four simultaneous leakages: global scaling
(estimation), global feature selection using full-data labels
(selection), 10\% test-row duplication (memorization), and
hyperparameter selection on the test set (selection). The compound
\(\Delta\)AUC inflation is d\(_z\) = 0.31. Compound effects are
sub-additive in 91.8\% of datasets: the median per-dataset ratio of
compound \(\Delta\)AUC to the sum of individual \(\Delta\)AUC values is
0.03. The dominant mechanism (selection) sets a ceiling; weaker
mechanisms contribute negligibly once selection leakage is present.
\textbf{Multi-class workflows.} Experiment J spans Classes I, II, and
III but not Class IV. Real workflows can simultaneously violate Class
III and Class IV, e.g., SMOTE-before-splitting on a temporally-ordered
medical dataset is both synthetic instance replication (Class III) and
partition-strategy mismatch (Class IV). The grammar's response column
lists one constraint per class; applying only the Class III remedy
(Constraint 3) leaves the temporal-boundary violation unaddressed, and
vice versa. The constraint-per-class table should be read as orthogonal
coverage, not mutually exclusive routing.

\textbf{Stack meta-leakage is null (Exp AK).} I predicted that
out-of-fold (OOF) stacking would introduce Class II leakage (prediction:
d\(_z\) \textgreater{} 0.3). The observed effect is d\(_z\) = -0.22,
negative, indicating OOF stacking is slightly conservative, not leaky.
This is the only failed prediction (Section~\ref{sec-scorecard}). The
OOF design (where each base model's predictions are generated without
seeing the target fold) successfully prevents meta-leakage. The result
validates stacking as a safe composition method.

\subsection{N-scaling separates Classes
I--III}\label{n-scaling-separates-classes-iiii}

Experiment AN measures the effect size at eight subsample sizes (n = 50,
100, 200, 500, 1,000, 2,000, 5,000, 10,000) across 493 datasets.
Datasets with \(\geq\) 2,000 rows (239) are tested at n = 50--2,000; the
152 datasets with \(\geq\) 10,000 rows are additionally tested at n =
5,000 and 10,000. Only datasets that succeed at all n-levels within each
tier are included (intersection set), trading censorship bias for
survivorship bias.

\begin{longtable}[]{@{}rcccc@{}}
\caption{Mean \(\Delta\)AUC at each subsample size. Rows n = 50--2,000:
239 datasets with \(\geq\) 2,000 rows (206 for oversampling). Rows n =
5,000--10,000: 152 datasets with \(\geq\) 10,000 rows; oversampling
omitted at extension scale (only \textasciitilde59 of 152 datasets meet
the class-imbalance criterion, a non-comparable population). Class II
persists at extension scale; Class I and III decline in the main
range.}\tabularnewline
\toprule\noalign{}
\(n\) & I: Estimation & II: Peeking & II: Seed & III: Oversample \\
\midrule\noalign{}
\endfirsthead
\toprule\noalign{}
\(n\) & I: Estimation & II: Peeking & II: Seed & III: Oversample \\
\midrule\noalign{}
\endhead
\bottomrule\noalign{}
\endlastfoot
50 & +0.005 & +0.115 & +0.137 & +0.247 \\
100 & +0.003 & +0.083 & +0.105 & +0.212 \\
200 & +0.002 & +0.068 & +0.084 & +0.177 \\
500 & +0.000 & +0.049 & +0.060 & +0.109 \\
1,000 & +0.001 & +0.048 & +0.056 & +0.079 \\
2,000 & +0.001 & +0.053 & +0.059 & +0.069 \\
5,000 & +0.000 & +0.036 & +0.007 & --- \\
10,000 & +0.000 & +0.039 & +0.005 & --- \\
\end{longtable}

\begin{figure}

\centering{

\includegraphics[width=1\linewidth,height=\textheight,keepaspectratio]{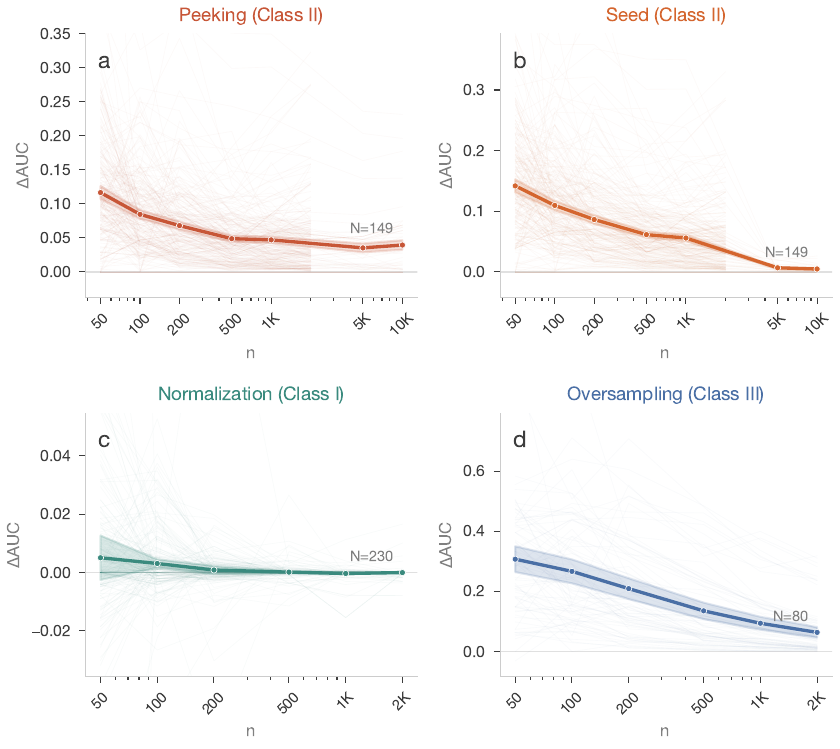}

}

\caption{\label{fig-nscaling}N-scaling separates the leakage classes.
(a,b) Class II extends to \(n = 10{,}000\): peeking retains a diversity
residual; seed decays to near-zero (pure noise exploitation). (c,d)
Class I and III shown at \(n = 50\)--\(2{,}000\) only: normalization is
already zero by \(n = 200\); oversampling declines steeply but extension
data is excluded due to survivorship bias (\(N\) drops from 149 to 59 in
the imbalanced subset). Thin lines = individual datasets; thick line =
mean. Shaded band = interquartile range.}

\end{figure}%

Four distinct regimes emerge:

\begin{enumerate}
\def\labelenumi{\arabic{enumi}.}
\item
  \textbf{Class I vanishes.} Normalization leakage is below +0.005 AUC
  at n = 50 and converges to near zero at n \(\geq\) 200. The O(p/n)
  bias term produces no detectable signal at practical dataset sizes.
\item
  \textbf{Class II persists, but mechanism-dependently.} Peeking (model
  selection on test-set performance) and seed inflation (best-of-10
  random seeds) both decay from n = 50 to n \(\approx\) 1,000, then
  differ at large n.~Peeking levels at a non-zero asymptotic floor (a
  diversity residual). Seed inflation empirically approaches zero at n
  \(\geq\) 5,000; the original prediction that seed retains a non-zero
  floor at large n was falsified (§sec-scorecard, AN-2s;
  Section~\ref{sec-scorecard}). At n = 2,000, peeking is +0.053 and seed
  is +0.059; at n = 10,000 on 152 larger datasets, peeking is +0.0393
  and seed is +0.0050. Peeking does not self-correct at large n; seed
  effectively does.

  Both mechanisms share the same mathematical root: selecting the best
  of \(K\) evaluations on the same holdout set. Peeking selects the best
  of \(K\) model configurations; seed inflation selects the best of
  \(K\) random seeds. In both cases the overshoot grows logarithmically
  in \(K\), and correlated draws (e.g., evaluations of related
  algorithms on the same folds, see Exp AQ) produce a smaller effective
  \(K\) and correspondingly smaller overshoot. The overshoot depends on
  the spread of scores (\(\sigma\)), not on the sample size. More data
  makes scores more stable (smaller \(\sigma\)), but it also makes the
  remaining spread matter more, so the search bonus stays a non-trivial
  fraction of whatever variance is left.
\item
  \textbf{Class III declines steeply.} The oversampling procedure is
  identical at every n; stratified subsampling preserves the class
  ratio, so the same fraction of synthetic rows is added whether n = 50
  or n = 2,000, but what changes is how much the model relies on them.
  At small n, the duplicated rows are a substantial fraction of the
  training signal; at large n, the model already generalizes well from
  legitimate data, and the leaked rows add little. Inflation at n =
  2,000 (+0.069) is 3.6 \(\times\) smaller than at n = 50 (+0.247).
  Extension to n = 5,000--10,000 is not shown: only 59 of the 152
  extension datasets have sufficient class imbalance for oversampling,
  and this survivorship produces a non-comparable pool.
\end{enumerate}

\textbf{Implication.} At the corpus median (\(n \approx 1{,}900\)),
peeking inflates by +0.040 AUC and seed by +0.045, both substantial.
Seed inflation vanishes by \(n = 5{,}000\) (100\% noise exploitation);
peeking retains a diversity residual at \(n = 100{,}000\). A structural
constraint that prevents repeated assessment on the same holdout data is
necessary at practical dataset sizes and wherever the i.i.d. assumption
is violated.

\subsection{Cross-validation confidence intervals are
miscalibrated}\label{sec-cv-coverage}

Bates, Hastie, and Tibshirani
(\citeproc{ref-bates2024crossvalidation}{2024}) showed in simulation
that standard k-fold confidence intervals can have coverage far below
the nominal level, with miscoverage two to three times the desired rate.
Experiment AO measures the magnitude of this gap empirically: the actual
coverage of nominal 95\% confidence intervals constructed from k-fold
cross-validation standard errors (N = up to 1,850 datasets per
algorithm, varying by convergence; the grand-mean 55\% coverage
statistic is computed over the 1,833 datasets where all three algorithms
(LR, RF, DT) converged). For each dataset, the ground-truth AUC is
estimated from a separate large held-out set (50\% of rows, withheld
before any CV) using a fixed random seed (42) for the held-out split per
dataset; coverage is the fraction of datasets whose CV-derived CI
contains this held-out estimate. Phase 1 (the 55\% grand-mean coverage
figure across LR, RF, DT) uses a single repetition of 5-fold CV per
dataset; Phase 2 (six-method comparison) uses three repetitions.
Differences between phases reflect this repetition count, not corpus or
seed changes.

\begin{longtable}[]{@{}lccc@{}}
\toprule\noalign{}
Algorithm & z-coverage & t-coverage & N \\
\midrule\noalign{}
\endhead
\bottomrule\noalign{}
\endlastfoot
LR & 56.3\% & 71.9\% & 1,833 \\
RF & 54.6\% & 69.9\% & 1,774 \\
DT & 54.5\% & 69.3\% & 1,850 \\
\end{longtable}

\begin{figure}

\centering{

\includegraphics[width=0.8\linewidth,height=\textheight,keepaspectratio]{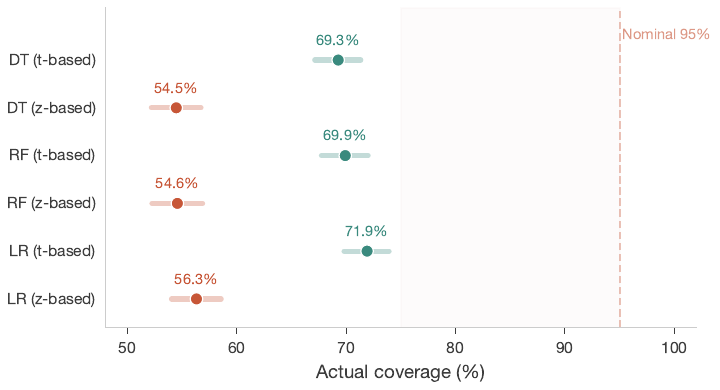}

}

\caption{\label{fig-coverage}Actual coverage of nominal 95\% confidence
intervals across three algorithms and six CI construction methods.
Dashed line = nominal 95\%. All methods fall short; Conservative-Z (M5)
performs best.}

\end{figure}%

A nominal 95\% z-based CI (the de facto default that fits the standard
error of the CV-mean across \(k = 5\) folds as
\(\mathrm{SD}_{\text{folds}} / \sqrt{k}\) and applies
\(z_{0.025} \approx 1.96\) as the critical value) achieves only 55\%
actual coverage (Wilson 95\% CI {[}53.8\%, 56.4\%{]} across \(N\) =
5,457 (algorithm, dataset) coverage observations; the interval treats
observations as independent, which slightly underestimates the true CI
because per-dataset algorithm coverages are not strictly independent).
The t-based correction (which replaces \(z\) with \(t_{k-1}\) but keeps
the same variance estimate) improves coverage to 70.4\%. The gap is
closeable: the Phase 2 comparison below shows Conservative-Z (M5)
reaches 87.4\% (within approximately 10 percentage points of nominal) by
using the fold SD directly, without dividing by \(\sqrt{k}\).

This result connects directly to Bengio and Grandvalet
(\citeproc{ref-bengio2004noUnbiased}{2004})'s impossibility theorem. The
fold-level standard error treats folds as independent, but folds share
training data (each point appears in k−1 of k folds). The resulting
underestimate of variance produces confidence intervals that are too
narrow by a factor of approximately 1.7 \(\times\).

\textbf{Six CI methods compared (Phase 2, N = 1,761 datasets).} To
identify whether better CI construction methods can recover the missing
coverage, I evaluated six approaches on two algorithms (LR, DT) with 3
repetitions of 5-fold CV per dataset:

\begin{longtable}[]{@{}
  >{\raggedright\arraybackslash}p{(\linewidth - 6\tabcolsep) * \real{0.2800}}
  >{\centering\arraybackslash}p{(\linewidth - 6\tabcolsep) * \real{0.1000}}
  >{\centering\arraybackslash}p{(\linewidth - 6\tabcolsep) * \real{0.1000}}
  >{\raggedright\arraybackslash}p{(\linewidth - 6\tabcolsep) * \real{0.5200}}@{}}
\caption{Six CI construction methods compared on LR and
DT.}\tabularnewline
\toprule\noalign{}
\begin{minipage}[b]{\linewidth}\raggedright
Method
\end{minipage} & \begin{minipage}[b]{\linewidth}\centering
LR
\end{minipage} & \begin{minipage}[b]{\linewidth}\centering
DT
\end{minipage} & \begin{minipage}[b]{\linewidth}\raggedright
Description
\end{minipage} \\
\midrule\noalign{}
\endfirsthead
\toprule\noalign{}
\begin{minipage}[b]{\linewidth}\raggedright
Method
\end{minipage} & \begin{minipage}[b]{\linewidth}\centering
LR
\end{minipage} & \begin{minipage}[b]{\linewidth}\centering
DT
\end{minipage} & \begin{minipage}[b]{\linewidth}\raggedright
Description
\end{minipage} \\
\midrule\noalign{}
\endhead
\bottomrule\noalign{}
\endlastfoot
M1: Naive z & 61.4\% & 57.2\% & Standard \(z_{0.025}\) on fold SEs \\
M2: t-corrected & 74.6\% & 71.5\% & \(t_{k-1}\) instead of \(z\) \\
M3: Nadeau-Bengio & 75.9\% & 75.5\% & Nadeau and Bengio
(\citeproc{ref-nadeau2003inference}{2003}) variance correction for
overlapping folds \\
M4: Bootstrap (B=500) & 66.9\% & \textbf{22.4\%} & Full k-fold CV rerun
on B bootstrap samples \\
\textbf{M5: Conservative-Z} & \textbf{87.4\%} & \textbf{86.5\%} & Fold
SD without \(\div\sqrt{k}\) \\
M6: Corrected Resampled-T & 75.9\% & 75.5\% & NB variance +
Bouckaert-Frank \(t\) \\
\end{longtable}

The best method, Conservative-Z (M5), uses the fold standard deviation
\emph{directly}, without dividing by \(\sqrt{k}\). This treats the \(k\)
folds as if they were the unit of replication rather than as a
sample-mean, which is the right conservative move when fold errors are
correlated (Bengio's impossibility result rules out an unbiased variance
correction; M5 deliberately over-widens instead of trying to estimate
the true variance precisely). M5 reaches approximately 87.4\% actual
coverage (within about 10 percentage points of nominal) and is the
recommended interval for practitioners reporting CV-based per-dataset
performance estimates.

\textbf{Scope of this recommendation.} The AO / M5 conversation concerns
\emph{per-dataset} CIs constructed from CV-fold SEs, the interval a
practitioner reports when they say ``this model achieves AUC = 0.85
{[}0.82, 0.88{]} on this dataset.'' The CIs reported elsewhere in this
paper (the \(d_z\) and \(\Delta\)AUC intervals defined in
Section~\ref{sec-effect-size}, computed as paired-difference
\(t\)-intervals across the 2,047-dataset corpus with SE =
\(s_\Delta / \sqrt{N}\)) are a different statistical quantity. Each
per-dataset \(\Delta\)AUC is CV-derived, but the corpus-level SE uses
between-dataset variance (\(s_\Delta / \sqrt{N}\)), not the per-dataset
fold-SD machinery that the AO experiment audits. The 55\%-vs-87\%
coverage gap therefore does not affect this paper's reported \(d_z\) and
\(\Delta\)AUC intervals; it changes the recommendation for \emph{future}
per-dataset CV reporting.

Three findings deserve attention. First, \textbf{bootstrap is
catastrophically anti-conservative for decision trees} (22.4\%
coverage). The instability of trees means that resampling produces
high-variance models whose fold-level CIs are far too narrow. Second, M3
and M6 produce identical coverage: both use the same variance estimate
with different critical values that happen to cancel. Third, M5 achieves
near-equal coverage for both stable (LR: 87.4\%) and unstable (DT:
86.5\%) algorithms.

\subsection{Prediction scorecard}\label{sec-scorecard}

For 13 experiments, I recorded directional predictions before data
collection. Ten are confirmed, two falsified, one qualified:

\begin{longtable}[]{@{}
  >{\raggedright\arraybackslash}p{(\linewidth - 6\tabcolsep) * \real{0.2353}}
  >{\raggedright\arraybackslash}p{(\linewidth - 6\tabcolsep) * \real{0.2353}}
  >{\raggedright\arraybackslash}p{(\linewidth - 6\tabcolsep) * \real{0.2353}}
  >{\centering\arraybackslash}p{(\linewidth - 6\tabcolsep) * \real{0.2941}}@{}}
\toprule\noalign{}
\begin{minipage}[b]{\linewidth}\raggedright
ID
\end{minipage} & \begin{minipage}[b]{\linewidth}\raggedright
Prediction
\end{minipage} & \begin{minipage}[b]{\linewidth}\raggedright
Evidence
\end{minipage} & \begin{minipage}[b]{\linewidth}\centering
Result
\end{minipage} \\
\midrule\noalign{}
\endhead
\bottomrule\noalign{}
\endlastfoot
AB & PCA: Class I, d\(_z\) \textless{} 0.1 & d\(_z\) = +0.08 & PASS \\
AF & Calibration: Class I, d\(_z\) \textless{} 0.15 & d\(_z\) = +0.05 &
PASS \\
AK & Stack: Class II, d\(_z\) \textgreater{} 0.3 & d\(_z\) = -0.22 &
\textbf{FAIL} \\
AQ & Screen: Class II, d\(_z\) = 0.1--0.5 & d\(_z\) = +0.27 & PASS \\
BA & SMOTE \(\approx\) random & diff \(\approx\) 0.01 & PASS \\
AN-1 & Class I \(\Delta\)AUC \(\to\) 0 at large n & +0.001 at n=2000 &
PASS \\
AN-2p & Class II peeking floor \textgreater{} 0 & c = 0.046, 95\% CI
{[}0.036, 0.052{]}. PASS on observable (\(c > 0\)), but decomposition
shows the residual at \(n = 10{,}000\) is \(\sim\!90\%\) algorithm
diversity, not persistent leakage (§ Discussion) & PASS\(^*\) \\
AN-2s & Class II seed floor \textgreater{} 0 & Floor model fit to
\(n \leq 2{,}000\) passes (CI excludes zero), but empirical values at
\(n = 5{,}000\) (+0.0073) and \(n = 10{,}000\) (+0.0050) approach zero &
FAIL \\
AP-1 & RF inflation \(\sim\) log(K) & R\(^2\) \textgreater{} 0.99 &
PASS \\
AP-2 & LR deterministic (sd \(\approx\) 0) & sd = 0.000 & PASS \\
AP-3 & RF inflation \textgreater{} LR inflation & RF \textgreater{} LR
at all K & PASS \\
AO-1 & z-coverage \textless{} 80\% & 55.1\% (N \(\approx\) 1,833) &
PASS \\
AO-2 & t-coverage \textless{} 90\% & 70.4\% (N \(\approx\) 1,833) &
PASS \\
\end{longtable}

Two falsified predictions are informative. AK (stack meta-leakage): I
predicted OOF stacking would leak; it does not: the OOF design prevents
the predicted information transfer. AN-2s (seed floor): the floor model
fit to \(n \leq 2{,}000\) excludes zero, but empirical values at
\(n \geq 5{,}000\) approach zero; seed inflation is 100\% noise
exploitation and vanishes at large \(n\), unlike peeking which retains a
diversity residual. The taxonomy is falsifiable: it makes strong
predictions, and when predictions fail, the failures have clear
mechanistic explanations.

\subsection{Updated taxonomy}\label{updated-taxonomy}

The experiments suggest a taxonomy organized by causal mechanism.
Classes I--III emerge from the twenty-eight core experiments (random CV
on iid benchmarks); Class IV emerges from the boundary experiment on 129
temporal datasets. This classification was developed from the data (not
pre-registered) and should be understood as a proposed organizing
framework, not a confirmed causal structure:

\begin{longtable}[]{@{}
  >{\centering\arraybackslash}p{(\linewidth - 12\tabcolsep) * \real{0.0600}}
  >{\raggedright\arraybackslash}p{(\linewidth - 12\tabcolsep) * \real{0.1400}}
  >{\raggedright\arraybackslash}p{(\linewidth - 12\tabcolsep) * \real{0.2600}}
  >{\raggedright\arraybackslash}p{(\linewidth - 12\tabcolsep) * \real{0.1600}}
  >{\centering\arraybackslash}p{(\linewidth - 12\tabcolsep) * \real{0.0900}}
  >{\centering\arraybackslash}p{(\linewidth - 12\tabcolsep) * \real{0.1200}}
  >{\raggedright\arraybackslash}p{(\linewidth - 12\tabcolsep) * \real{0.1700}}@{}}
\caption{Updated taxonomy organized by causal
mechanism.}\label{tbl-taxonomy}\tabularnewline
\toprule\noalign{}
\begin{minipage}[b]{\linewidth}\centering
Class
\end{minipage} & \begin{minipage}[b]{\linewidth}\raggedright
Name
\end{minipage} & \begin{minipage}[b]{\linewidth}\raggedright
Mechanism
\end{minipage} & \begin{minipage}[b]{\linewidth}\raggedright
Experiments
\end{minipage} & \begin{minipage}[b]{\linewidth}\centering
d\(_z\)
\end{minipage} & \begin{minipage}[b]{\linewidth}\centering
\(\Delta\)AUC
\end{minipage} & \begin{minipage}[b]{\linewidth}\raggedright
n-scaling
\end{minipage} \\
\midrule\noalign{}
\endfirsthead
\toprule\noalign{}
\begin{minipage}[b]{\linewidth}\centering
Class
\end{minipage} & \begin{minipage}[b]{\linewidth}\raggedright
Name
\end{minipage} & \begin{minipage}[b]{\linewidth}\raggedright
Mechanism
\end{minipage} & \begin{minipage}[b]{\linewidth}\raggedright
Experiments
\end{minipage} & \begin{minipage}[b]{\linewidth}\centering
d\(_z\)
\end{minipage} & \begin{minipage}[b]{\linewidth}\centering
\(\Delta\)AUC
\end{minipage} & \begin{minipage}[b]{\linewidth}\raggedright
n-scaling
\end{minipage} \\
\midrule\noalign{}
\endhead
\bottomrule\noalign{}
\endlastfoot
I & Estimation & Parameter averaging O(p/n) & A, A2, D, E, F, T, Q, CE,
AB, AF & \(\approx 0\) & \(\approx 0\) & Vanishes \\
II & Selection & Label/test info selects model & B, K, BB, AI, AQ, AC,
AP & 0.27--0.93 & +0.013--0.045 & grows with \(\log K\); decays at large
\(n\) \\
III & Memorization & Training on evaluation data & G, H, BA & 0.29--1.38
& +0.001--0.073 & f(capacity, fraction) \\
IV & Boundary & Partition strategy mismatches deployment boundary & P,
Boundary exp. & --- & +0.01 (group\(^g\)), +0.023\(^b\) (temporal) &
f(non-stationarity) \\
\end{longtable}

\(^g\)Experiment P: random CV vs GroupKFold on a click-prediction
dataset with \texttt{ad\_id} as group column. Most public benchmark
datasets strip ID columns at upload, limiting the scope
(Section~\ref{sec-boundary}). \(^b\)Mean pure temporal effect across 14
non-FOREX datasets with verified genuine timestamps; d\(_z\) not
reported (different experimental design from Classes I--III). Near zero
on benchmarks without real drift. Domain-dependent, not universal
(Section~\ref{sec-boundary}).

\subsection{Internal Validation}\label{internal-validation}

Before analysis, each dataset is assigned by content hash to a
\emph{discovery} or \emph{confirmation} split (50/50). All testable
effects are confirmed on the held-out confirmation split with zero
failures. The strongest effects show near-identical magnitudes across
splits: peeking (discovery d\(_z\) = 0.94, confirmation d\(_z\) = 0.92),
seed inflation (discovery \(\Delta\)AUC = 0.046, confirmation
\(\Delta\)AUC = 0.044), duplication (discovery d\(_z\) = 0.90,
confirmation d\(_z\) = 0.89). The zero-failure rate is itself a finding:
these are stable, reproducible phenomena, not artifacts of specific
dataset subsets.

\section{Discussion}\label{discussion}

Every ML textbook warns: normalize inside the fold. That advice is
correct. Nine experiments confirm the magnitude is negligible at typical
dataset sizes (\(\Delta\)AUC \textless{} 0.005 in all cases).
Scikit-learn Pipelines (\citeproc{ref-pedregosa2011scikit}{Pedregosa et
al. 2011}) already handle this correctly: a Pipeline with a
StandardScaler first step learns scaling parameters from the training
portion of each fold and applies the same parameters to the held-out
portion. The tooling solution for Class I leakage exists and works. The
problem is that the pedagogical emphasis on per-fold preprocessing is
disproportionate to the risk it mitigates.

The results suggest a different ordering. The leakage types that
actually inflate performance estimates are: (1) selection leakage,
particularly peeking (\(\Delta\)AUC = +0.040) and seed cherry-picking
(\(\Delta\)AUC = +0.045); (2) memorization leakage, particularly in
high-capacity models (\(\Delta\)AUC = +0.013--0.026 at 10\%
duplication); and (3) early stopping on test data (d\(_z\) = 0.46).
Within the iid tabular regime studied here, practitioners should audit
for these first; for temporal, grouped, or spatial data the boundary
experiment suggests Class IV may dominate and the priority ordering
changes accordingly (see Limitation 13).

These measured effects are lower bounds: random stratified CV assumes
exchangeability across rows, and on data with group, temporal, or
spatial structure the ``clean'' baseline silently absorbs structural
contamination (\citeproc{ref-roberts2017cross}{Roberts et al. 2017};
\citeproc{ref-valavi2019blockcv}{Valavi et al. 2019}). The boundary
experiment (Section~\ref{sec-boundary}) quantifies this directly.

\subsection{Selection leakage decomposes into noise exploitation and
genuine
diversity}\label{selection-leakage-decomposes-into-noise-exploitation-and-genuine-diversity}

Selection leakage via model-selection peeking is the most universal
threat in the landscape at practical dataset sizes (d\(_z\) = 0.93,
\(\Delta\)AUC = +0.040, 92\% prevalence). It is also the hardest to
detect. The non-monotonicity at \(k = 1\) means a practitioner who
evaluates only a few configurations on the test set may observe no
inflation and conclude the procedure is safe.

Every selection mechanism decomposes into two components:
\(\Delta = \Delta_{\text{diversity}} + \Delta_{\text{noise}}\), where
the diversity term reflects genuine performance differences across the
selection pool and the noise term reflects exploitation of estimation
variance. The noise term grows logarithmically in \(K\) (measured
directly in the dose-response below, \(R^2 > 0.99\)), with the AUC
standard error \(\sigma \sim 1/\sqrt{n}\) setting its scale.

The seed experiment provides a direct empirical estimate of the noise
component, because seed inflation has zero diversity (all seeds estimate
the same underlying AUC):
\(\hat{\sigma} = \Delta_{\text{seed}} / g(K_{\text{seed}})\). The
decomposition below uses \(K_{\text{seed}} = K_{\text{peek}} = 10\) (Exp
AI best-of-10, Exp B at \(k = 10\)); \(g(K)\) is the logarithmic scaling
defined above (modelling assumption; see Limitation 11). Subtracting the
noise prediction from the peeking effect yields the diversity component
at each sample size:

\begin{longtable}[]{@{}
  >{\centering\arraybackslash}p{(\linewidth - 10\tabcolsep) * \real{0.1300}}
  >{\centering\arraybackslash}p{(\linewidth - 10\tabcolsep) * \real{0.1400}}
  >{\centering\arraybackslash}p{(\linewidth - 10\tabcolsep) * \real{0.1400}}
  >{\centering\arraybackslash}p{(\linewidth - 10\tabcolsep) * \real{0.1600}}
  >{\centering\arraybackslash}p{(\linewidth - 10\tabcolsep) * \real{0.1600}}
  >{\centering\arraybackslash}p{(\linewidth - 10\tabcolsep) * \real{0.1400}}@{}}
\caption{Noise/diversity decomposition. Values stored in
\texttt{claims.json} under \texttt{decomp.*}; cohort sizes vary across
\(n\)-tiers due to subsampling feasibility (datasets with \(<\!n\) rows
are excluded from each tier).}\tabularnewline
\toprule\noalign{}
\begin{minipage}[b]{\linewidth}\centering
\(n\)
\end{minipage} & \begin{minipage}[b]{\linewidth}\centering
Peeking \(\Delta\)
\end{minipage} & \begin{minipage}[b]{\linewidth}\centering
Seed \(\Delta\)
\end{minipage} & \begin{minipage}[b]{\linewidth}\centering
Noise fraction
\end{minipage} & \begin{minipage}[b]{\linewidth}\centering
Diversity fraction
\end{minipage} & \begin{minipage}[b]{\linewidth}\centering
\(N_{\text{datasets}}\)
\end{minipage} \\
\midrule\noalign{}
\endfirsthead
\toprule\noalign{}
\begin{minipage}[b]{\linewidth}\centering
\(n\)
\end{minipage} & \begin{minipage}[b]{\linewidth}\centering
Peeking \(\Delta\)
\end{minipage} & \begin{minipage}[b]{\linewidth}\centering
Seed \(\Delta\)
\end{minipage} & \begin{minipage}[b]{\linewidth}\centering
Noise fraction
\end{minipage} & \begin{minipage}[b]{\linewidth}\centering
Diversity fraction
\end{minipage} & \begin{minipage}[b]{\linewidth}\centering
\(N_{\text{datasets}}\)
\end{minipage} \\
\midrule\noalign{}
\endhead
\bottomrule\noalign{}
\endlastfoot
50 & 0.126 & 0.152 & 95\% & 5\% & 232 \\
200 & 0.074 & 0.094 & 100\% & 0\% & 239 \\
1,000 & 0.050 & 0.059 & 94\% & 6\% & 239 \\
2,000 & 0.053 & 0.059 & 88\% & 12\% & 239 \\
5,000 & 0.036 & 0.007 & 16\% & 84\% & 151 \\
10,000 & 0.039 & 0.005 & 10\% & 90\% & 152 \\
50,000 & 0.037 & 0.004 & 9\% & 91\% & 94 \\
100,000 & 0.035 & 0.003 & 8\% & 92\% & 94 \\
\end{longtable}

At the corpus median (\(n \approx 1{,}900\)), peeking is consistent with
an approximately 90\% noise-exploitation share, real selection bias that
inflates the reported score beyond the true generalization performance,
given the modelling assumption that seed inflation measures pure noise
(Limitation 11). At \(n = 100{,}000\) (94 datasets), the noise share has
fully decayed to \textasciitilde8\% on the same decomposition, and the
residual \(\Delta \approx 0.035\) reflects genuine algorithm diversity:
random forest truly outperforms logistic regression on these datasets,
and that gap does not shrink with sample size. The diversity component
is not leakage; it is what model selection is for. Supporting evidence:
at \(n \leq 2{,}000\), peeking and seed are correlated at \(r > 0.96\)
across datasets (both driven by the same \(\sigma\)); at
\(n = 10{,}000\), the correlation collapses to \(r = -0.06\); the noise
component that linked them has vanished. The convergence continues
smoothly to \(n = 100{,}000\), where seed inflation is +0.003 AUC
(effectively zero).

Seed inflation at the corpus median (\(\Delta\)AUC = +0.045, d\(_z\) =
0.89) is as large as peeking but decays to +0.0050 at \(n = 10{,}000\).
This confirms the first-principles prediction: seed inflation is 100\%
noise exploitation with zero diversity (all seeds estimate the same
underlying AUC).

\subsection{Seed cherry-picking: pure noise
exploitation}\label{seed-cherry-picking-pure-noise-exploitation}

Seed inflation receives far less attention than preprocessing leakage,
despite being mentioned by Lones
(\citeproc{ref-lones2024pitfalls}{2024}) and falling under the broader
umbrella of researcher degrees of freedom. The logarithmic dose-response
(R\(^2\) \textgreater{} 0.99) means cherry-picking scales predictably
with effort: a researcher who tries 10 seeds and reports the best
inflates by +0.016 AUC on average; 100 seeds inflates by +0.026. Cawley
and Talbot (\citeproc{ref-cawley2010overfitting}{2010}) analyzed the
model-selection overfitting this reflects. In the adaptive data analysis
framework of Dwork et al. (\citeproc{ref-dwork2015reusable}{2015}), each
seed evaluation is an adaptive query against the holdout.

At the corpus median (\(n \approx 1{,}900\)), the per-dataset effect of
+0.016 AUC from a single reporting decision is real bias. But because
seed inflation is 100\% noise exploitation (zero diversity: all seeds
estimate the same underlying AUC), it decays as \(1/\sqrt{n}\) and
effectively vanishes by \(n = 5{,}000\). This makes it qualitatively
different from peeking, where the diversity component persists.

\subsection{Boundary effects: the leakage that random CV
hides}\label{sec-boundary}

The 28 core experiments above all use random stratified
cross-validation. This is the standard protocol, and the standard
protocol has a blind spot. When data has temporal ordering, group
structure, or spatial proximity, random CV scatters correlated
observations across folds, and the ``clean'' baseline absorbs structural
contamination silently. The measured selection effects are what remains
on top of this potentially already-leaked baseline.

A boundary experiment tests the magnitude directly: for each dataset
with a genuine temporal column, compare the AUC from random 5-fold CV
against the AUC from walk-forward evaluation respecting the temporal
ordering. To separate temporal leakage from the training-set-size
confound (walk-forward trains on smaller early folds), I also compute a
size-matched random control with the same expanding-window fold sizes
but shuffled row order. The pure temporal effect is the gap between the
size-matched random and walk-forward evaluations.

I scanned 1,853 cached OpenML datasets for temporal columns, identified
129 with time-related column names, and verified 14 non-financial
datasets with genuine timestamps after filtering out false positives
(columns named \texttt{hours-per-week}, \texttt{wage\_per\_hour},
\texttt{time\_in\_hospital} are durations, not temporal ordering). An
additional 92 FOREX currency-pair datasets serve as a null control:
efficient market prices have no exploitable temporal structure.
\textbar{} Condition \textbar{} \(N\) \textbar{} Pure temporal
\(\Delta\)AUC \textbar{} Positive fraction \textbar{}
\textbar:---\textbar:---:\textbar:---:\textbar:---:\textbar{} \textbar{}
FOREX (null control) \textbar{} 92 \textbar{} \(-0.006\) \textbar{} 42\%
\textbar{} \textbar{} Non-FOREX with genuine timestamps \textbar{} 14
\textbar{} \(+0.023\) \textbar{} 57\% \textbar{} \textbar{} Non-FOREX
with spurious time columns \textbar{} 23 \textbar{} \(+0.002\)
\textbar{} --- \textbar{} : Boundary experiment results.
\{tbl-colwidths=``{[}40,10,25,25{]}''\}

The FOREX null confirms the instrument: random and walk-forward
evaluation agree on efficient market data. The 14 genuine temporal
datasets show a mean pure temporal effect of +0.023, driven by domains
with real concept drift: credit card fraud (+0.12 on 20K subsample,
+0.013 on full 284K after size control), electricity market (+0.03),
drug directory (+0.03), road safety (+0.02). The spurious time columns
(+0.002) confirm that the heuristic correctly separates real temporal
structure from column-name artifacts.

The boundary effect is domain-dependent, not universal. On typical
benchmark datasets without temporal structure, random CV is adequate.
Where real distribution drift exists (fraud patterns that evolve, sensor
responses that degrade, markets that shift regimes), the effect is
substantial and invisible under the standard protocol. \textbf{Feature
selection leakage scales with dimensionality.} On datasets where
\(p < n\) (typical benchmarks), wrapper feature selection before
splitting is negligible (+0.001 mean across 72 datasets). On
high-dimensional datasets where \(p/n >\) 0.1 (genomics, proteomics,
text), the effect becomes substantial: +0.018 mean across 49 datasets,
with individual effects up to +0.10 on datasets with \(p \approx n\).
This confirms Ambroise and McLachlan
(\citeproc{ref-ambroise2002selection}{2002})'s finding on gene
expression data and extends it to a broader corpus. \textbf{Metric
selection flips model rankings.} On 100 datasets, comparing LR and RF
across 6 standard metrics (AUC, F1, accuracy, precision, recall, MCC),
the choice of metric changes which model wins on 31\% of datasets. This
is not a pipeline error; it is a decision sensitivity that the
practitioner controls, and it operates on the same selection principle
as peeking: the researcher who reports the most flattering metric is
making a selection decision. \textbf{Group leakage.} Click prediction
with \texttt{ad\_id} as group column shows +0.01 AUC (random CV vs
GroupKFold). Most public benchmark datasets strip ID columns at upload,
limiting the scope of this experiment. Clinical and e-commerce datasets
with patient or customer IDs would show larger effects, but these
require data use agreements not available for this study.

\subsection{Bayesian meta-regression: mechanism matters more than
dataset}\label{sec-meta-regression}

To test whether raw correlations between leakage severity and dataset
characteristics (Section~\ref{sec-raw-correlations}) survive proper
hierarchical modeling, I fit a Bayesian measurement-error
meta-regression (PyMC 5.28 with numpyro backend, 4 chains \(\times\)
2,000 samples, target\_accept = 0.9).

\textbf{Model structure.} Three-level hierarchy: leakage class fixed
effects (\(\alpha_{\text{class}}\)), experiment random effects
(\(u_{\text{exp}}\), non-centered parameterization), and dataset random
effects (\(u_{\text{ds}}\), cross-experiment). Moderators: z-scored
log(n), log(p), and imbalance. Known within-study standard errors
(\(\text{se}_i\)) from paired repetitions. Region of Practical
Equivalence (ROPE) on the \(\Delta\)AUC scale \(= [-0.02, +0.02]\),
following Kruschke (\citeproc{ref-kruschke2018rejecting}{2018}) (effects
within this interval are treated as practically null). This ROPE was
fixed on substantive grounds \emph{before} the meta-regression was run:
0.02 sits below the per-dataset detection floor (\(d_z = 0.057\) at
\(N = 2{,}047\)) and below any clinically actionable threshold in the
contexts studied.

\textbf{Priors.} \(\alpha_{\text{class}} \sim \mathcal{N}(0, 0.1)\);
\(\beta_{\text{moderator}} \sim \mathcal{N}(0, 0.05)\);
\(\tau_{\text{exp}}, \tau_{\text{ds}} \sim \text{HalfNormal}(0.05)\);
\(\sigma_{\text{resid}} \sim \text{HalfNormal}(0.1)\). Priors are weakly
informative, centered on the expectation that most leakage effects are
small in AUC units. A sensitivity analysis with doubled prior widths
produces posterior means within 0.001 AUC of the primary analysis.

\textbf{Data.} 12,103 observations across 7 experiments (A:
normalization, B\(_{k=10}\): peeking, AI: seed inflation, AQ: screen
selection, BA: oversampling, BB: early stopping, H\(_{10\%}\):
duplication) and 2,047 datasets (below the
\(7 \times 2{,}047 = 14{,}329\) ceiling because some experiments
restrict to eligible-dataset subsets, e.g., BA oversampling to
imbalanced classes, and convergence failures are excluded per
experiment; per-experiment counts vary). Each observation is a (dataset,
experiment) pair with \(\Delta_i\) = mean leaky − mean clean across 5
repetitions, and \(\text{se}_i\) from the standard error of paired
differences.

\begin{longtable}[]{@{}
  >{\raggedright\arraybackslash}p{(\linewidth - 8\tabcolsep) * \real{0.1400}}
  >{\centering\arraybackslash}p{(\linewidth - 8\tabcolsep) * \real{0.1500}}
  >{\centering\arraybackslash}p{(\linewidth - 8\tabcolsep) * \real{0.2000}}
  >{\raggedright\arraybackslash}p{(\linewidth - 8\tabcolsep) * \real{0.1700}}
  >{\raggedright\arraybackslash}p{(\linewidth - 8\tabcolsep) * \real{0.3400}}@{}}
\caption{Bayesian meta-regression posteriors.}\tabularnewline
\toprule\noalign{}
\begin{minipage}[b]{\linewidth}\raggedright
Parameter
\end{minipage} & \begin{minipage}[b]{\linewidth}\centering
Posterior mean
\end{minipage} & \begin{minipage}[b]{\linewidth}\centering
94\% HDI\footnote{94\% HDI (highest density interval) is the Bayesian
  analogue of a frequentist confidence interval, following the default
  in ArviZ; the 1\% difference from a 95\% interval is negligible at
  this sample size.}
\end{minipage} & \begin{minipage}[b]{\linewidth}\raggedright
Classification
\end{minipage} & \begin{minipage}[b]{\linewidth}\raggedright
Meaning
\end{minipage} \\
\midrule\noalign{}
\endfirsthead
\toprule\noalign{}
\begin{minipage}[b]{\linewidth}\raggedright
Parameter
\end{minipage} & \begin{minipage}[b]{\linewidth}\centering
Posterior mean
\end{minipage} & \begin{minipage}[b]{\linewidth}\centering
94\% HDI{}
\end{minipage} & \begin{minipage}[b]{\linewidth}\raggedright
Classification
\end{minipage} & \begin{minipage}[b]{\linewidth}\raggedright
Meaning
\end{minipage} \\
\midrule\noalign{}
\endhead
\bottomrule\noalign{}
\endlastfoot
\(\alpha_{\text{Class I}}\) & -0.002 & {[}-0.018, 0.013{]} & NULL &
Class I (estimation) intercept \\
\(\alpha_{\text{Class II}}\) & 0.006 & {[}-0.010, 0.022{]} &
Inconclusive & Class II (selection) intercept \\
\(\alpha_{\text{Class III}}\) & 0.025 & {[}-0.004, 0.053{]} &
Inconclusive & Class III (memorization) intercept \\
\(\beta_{\log n}\) & -0.002 & {[}-0.003, -0.002{]} & NULL & \(\log(n)\)
moderator \\
\(\beta_{\log p}\) & 0.003 & {[}0.002, 0.003{]} & NULL & \(\log(p)\)
moderator \\
\(\beta_{\text{imbalance}}\) & -0.001 & {[}-0.001, 0.000{]} & NULL &
class-imbalance moderator \\
\(\tau_{\text{exp}}\) & 0.013 & --- & --- & between-experiment SD
(variance component) \\
\(\tau_{\text{ds}}\) & 0.005 & --- & --- & between-dataset SD (variance
component) \\
\(\sigma_{\text{resid}}\) & 0.025 & --- & --- & residual SD \\
\end{longtable}

\emph{MCMC convergence achieved (R-hat = 1.0, ESS\(_{\text{bulk}}\)
\textgreater{} 1,700; 30 divergences (0.4\%) clustered near the
\(\tau_{\text{ds}} \to 0\) boundary, posteriors unchanged by their
exclusion).}

\textbf{Variance-ratio finding (corroborative, not primary).} The
between-experiment variance (\(\tau_{\text{exp}}\) = 0.013) exceeds the
between-dataset variance (\(\tau_{\text{ds}}\) = 0.005) by a factor of
2.6\(\times\). This is \emph{consistent with} leakage \emph{mechanism}
as the dominant moderator of effect size, rather than dataset
characteristics, but the ratio is not an algorithm-free contrast
(Limitation 12), so the primary evidence for the mechanism-dominance
conclusion remains the raw correlations in
Section~\ref{sec-raw-correlations}, which are algorithm-stratified. Two
structural caveats also apply: (1) experiment type is partially
confounded with algorithm (e.g., DT appears only in Class III
experiments), so the variance ratio may partly reflect algorithm
differences; and (2) each experiment belongs to exactly one leakage
class, so \(\tau_{\text{exp}}\) absorbs between-class variance by
construction; the ratio confirms that the proposed classification
captures real variance, not that the classification itself is uniquely
correct.

\textbf{Note on the Class II and Class III ROPE classifications.} Both
\(\alpha_{\text{Class II}}\) and \(\alpha_{\text{Class III}}\) classify
as ``Inconclusive'' not because Class II/III effects are weak, but
because the meta-regression pools heterogeneous experiments within each
class into a single intercept. Partial pooling compresses these
intercepts toward the grand mean. The individual experiments remain
significant (Class II: d\(_z\) = 0.27--0.93 across selection mechanisms;
the lower bound, screen selection, is d\(_z\) = 0.27, an order of
magnitude above the d\(_z\) = 0.057 detection floor at \(N\) = 2,047.
Class III: d\(_z\) up to 1.38 for DT at 30\% duplication, with
\(N > 1{,}800\)). The ``Inconclusive'' classifications reflect
within-class heterogeneity absorbed by the hierarchical model, not doubt
about whether Class II or Class III effects exist.

\textbf{Why the Bayesian result reverses the raw correlations.} A naive
Spearman correlation between \(\log(n)\) and leakage severity is
negative within individual experiments (e.g., \(r =\) -0.27 for seed
inflation, \(p < 0.001\)). This looks like evidence that bigger datasets
leak less. The pattern is consistent with Simpson's paradox
(\citeproc{ref-simpson1951interpretation}{Simpson 1951}). The proposed
explanation: oversampling (Class III) produces the largest effects and
disproportionately affects smaller, imbalanced datasets; normalization
(Class I) produces near-zero effects across all sizes. Pool them, and
``big dataset'' becomes a proxy for ``not the oversampling experiment.''
The hierarchical model resolves this by giving each experiment its own
intercept. Once conditioned on experiment type, the \(\log(n)\) slope
collapses to \(\beta\) = -0.002, firmly null. Within any single
experiment, dataset size explains almost nothing.

\textbf{Implication:} Leakage severity cannot be predicted from dataset
features alone; within any single experiment, dataset characteristics
explain almost no additional variance once experiment type is
conditioned out (the three moderators
\(\beta_{\log n}, \beta_{\log p}, \beta_{\text{imbalance}}\) in the
meta-regression table above all classify as NULL in ROPE). There is no
``safe zone'' where leakage prevention can be relaxed. Prevention must
be unconditional: a structural property of the workflow, not a
conditional check on dataset characteristics.

\subsubsection{Raw moderator correlations
(non-hierarchical)}\label{sec-raw-correlations}

For completeness, the raw Spearman rank correlations across all 2,047
datasets:

\begin{longtable}[]{@{}lcccc@{}}
\caption{Spearman rank correlations across 2,047
datasets.}\label{tbl-raw-correlations}\tabularnewline
\toprule\noalign{}
Moderator & Oversample & Seed & Screen & Early stop \\
\midrule\noalign{}
\endfirsthead
\toprule\noalign{}
Moderator & Oversample & Seed & Screen & Early stop \\
\midrule\noalign{}
\endhead
\bottomrule\noalign{}
\endlastfoot
p/n ratio & +0.55 & +0.19 & +0.13 & +0.08 \\
log(n) & -0.37 & -0.27 & -0.13 & -0.08 \\
Imbalance & -0.14 & -0.18 & -0.04 & -0.08 \\
\end{longtable}

These correlations describe the marginal association across all
experiments pooled. They are reported here to make the Simpson's-paradox
motivation for the hierarchical model visible: once experiment type is
conditioned on (Section~\ref{sec-meta-regression}), the direction of
these correlations reverses or collapses to null. The hierarchical
analysis is the substantive test; the marginal table is the diagnostic
that justifies it.

\subsection{The memorization capacity principle}\label{sec-capacity}

The capacity ordering NB \(<\) LR \(<\) XGB \(<\) RF \(<\) KNN \(<\) DT
for memorization leakage holds at 10\% and 30\% duplication and supports
a general principle: susceptibility to memorization leakage tracks a
model's \emph{memorization capacity} (its ability to overfit individual
instances), distinct from predictive capacity. At 5\% duplication a
single inversion appears (KNN d\(_z\) = 0.92 exceeds DT d\(_z\) = 0.87):
at low duplication density, DT's regularization-free memorization
advantage has not yet activated; with sufficient duplicates the ordering
recovers. The mechanism at 30\% spans the algorithms: DT (d\(_z\) =
1.38) memorizes duplicates exactly once present; KNN (d\(_z\) = 1.25)
stores training instances directly; RF (d\(_z\) = 0.95) dilutes per-tree
memorization through bagging; XGBoost (d\(_z\) = 0.86) is regularized by
default (max\_depth = 6, L2 penalty); LR (d\(_z\) = 0.48) has no
capacity for instance memorization; NB (d\(_z\) = 0.42) is most
constrained by class-conditional independence. Neural networks trained
without explicit regularization should show effects comparable to or
exceeding RF at sufficient duplication density, a testable prediction.

\subsection{Implications for tooling
design}\label{implications-for-tooling-design}

The taxonomy has a direct engineering implication. A workflow API that
structurally prevents Class II (selection) and Class III (memorization)
leakage, not by documentation but by making the incorrect workflow
inexpressible, would eliminate the leakage types that actually matter.
Class I errors would also be rejected on principle, even though their
measured effect is negligible. The API would enforce four structural
constraints: assess once per holdout (preventing selection pressure on
test data), prepare after split and per fold (preventing preprocessing
leakage), type-safe transitions (preventing skipped or misordered
steps), and rejection of unregistered data at fit time (preventing
untagged data from bypassing the grammar). I develop this argument and
present a formal grammar for ML workflows (typed partitions, once-only
assessment gates, and per-fold preparation scoping) in a companion paper
(\citeproc{ref-roth2026grammar}{Roth 2026}).

\subsection{Limitations}\label{limitations}

This study has several limitations:

\begin{enumerate}
\def\labelenumi{\arabic{enumi}.}
\item
  \textbf{Scope.} I test binary classification with six primary
  algorithms (NB, LR, XGB, RF, KNN, DT). Multiclass, regression, and
  additional algorithms (neural networks) are out of scope, though
  secondary experiments (A2, K, AO) include additional algorithms.
\item
  \textbf{Synthetic leakage.} The experiments inject leakage
  synthetically via controlled perturbations. This is the right design
  for comparative magnitude ranking (which types are larger than which),
  because it isolates each mechanism. However, naturally occurring
  leakage (AutoML tools that internally reuse holdouts, feature
  engineering scripts run on the full dataset before the split is
  defined, leakage inherited from shared benchmarks) may produce
  different magnitudes due to interactions with dataset-specific
  structure, larger configuration spaces, and opaque preprocessing
  chains. The assumption that synthetic treatment injection is
  representative of natural leakage is not tested in this study.
\item
  \textbf{Internal scorecard validation, not adversarial
  pre-registration.} The confirmation split (discovery/confirmation
  halves) is an internal replication procedure: 10 of 13 directional
  predictions confirmed on the held-out half, with two failures (stack
  meta-leakage, seed floor) and one qualified result (peeking
  diversity), mechanistically informative (Section~\ref{sec-scorecard}).
  Both the scorecard registration and the experiments were designed by
  the same author. The 10/13 confirmation rate is consistent with the
  underlying mechanisms but does not constitute an adversarial
  pre-registration test. Independent replication on a corpus the author
  did not curate is the correct falsifiability check.
\item
  \textbf{N-scaling design.} The AN peeking sub-experiment uses a
  multi-algorithm leaky path vs single-algorithm clean path, so its
  absolute inflation is not directly comparable to Experiment B. AN
  measures how leakage scales with sample size; Experiment B (d\(_z\) =
  0.93) measures the effect size itself.
\item
  \textbf{sklearn-only implementation.} All experiments use
  scikit-learn. Different frameworks may produce slightly different
  magnitudes; causal mechanisms are framework-independent.
\item
  \textbf{Scope beyond iid.} The core experiments assume iid tabular
  data. The boundary experiment tests temporal leakage on 129 datasets,
  but group leakage is tested on only one dataset. Informative
  missingness and domain-knowledge feature leakage are outside scope.
  The practical recommendation to deprioritize estimation leakage should
  not be applied without domain-specific validation in clinical,
  temporal, or high-stakes contexts. The +0.023 mean temporal effect
  (genuine-timestamp subset, N=14) is a magnitude estimate on a small
  benchmark subset, \emph{not an upper bound}: production time-series
  workflows with lookahead-feature construction, rolling-window target
  leakage, or embargo violations can produce substantially larger
  inflation that this experiment cannot detect by design.
\item
  \textbf{Power law identification.} The 3-parameter floor model is fit
  to 6 data points (3 residual df). Profile likelihood 95\% CIs for the
  floor parameter exclude zero (peeking: {[}0.036, 0.052{]}; seed:
  {[}0.033, 0.057{]}), but the floor estimate should be interpreted as
  ``non-zero'' rather than as a precise asymptotic value. The empirical
  values at \(n = 10{,}000\) (+0.0393/+0.0050) are consistent with the
  fitted floors but the claim is bounded to the tested range. A
  parametric bootstrap (10,000 resamples over datasets) produces 95\%
  CIs of {[}0.035, 0.052{]} for peeking and {[}0.02, 0.055{]} for seed,
  consistent with the profile likelihood intervals and confirming the
  exclusion of zero.
\item
  \textbf{N-scaling variance confound.} Raw effects at small \(n\) are
  confounded with decreasing cross-validation error-bar width at larger
  sample sizes (\citeproc{ref-varoquaux2018crossvalidation}{Varoquaux
  2018}). From \(n = 50\) to \(n = 2{,}000\) the raw \(\Delta\)AUC decay
  tracks both the leakage signal and the shrinking estimator variance.
  However, the standardized effect size \(d_z\) remains above 0.96 at
  all sample sizes tested (range: 0.96--1.47 for peeking, 1.04--1.56 for
  seed), indicating that the leakage effect is real at every \(n\) and
  not an artifact of small-sample variance. The primary evidence for the
  non-zero floor is the empirical value at \(n = 10{,}000\) (where CV
  variance is negligible), not the curve shape at small \(n\). A 50\%
  hold-out estimate on the \(n = 10{,}000\) datasets (the same machinery
  used for the CV-coverage experiment, Section~\ref{sec-cv-coverage})
  would yield a near-population-level AUC ground truth, against which
  the peeking-vs-clean gap could be re-estimated independently of the
  CV-mean machinery, queued as future work.
\item
  \textbf{CV coverage aggregation.} The 55\% actual coverage at nominal
  95\% is a grand mean across 1,833 datasets. A size-stratified analysis
  reveals that the miscalibration is remarkably stable: coverage is 57\%
  for \(n < 200\), 56\% for \(200 \leq n < 500\), 56\% for
  \(500 \leq n < 1{,}000\), 57\% for \(1{,}000 \leq n < 5{,}000\), and
  56\% for \(n \geq 5{,}000\). This is a structural property of the CV
  variance estimator, not a small-sample artifact.
\item
  \textbf{Baseline is not an oracle.} The clean baseline is a correct
  5-fold cross-validation workflow, not a true held-out generalization
  estimate. As noted in Section~\ref{sec-meta-regression} and the design
  section, on datasets with group, temporal, or spatial structure the
  clean baseline itself already absorbs structural contamination because
  random CV scatters correlated observations across folds. The reported
  \(\Delta\)AUC values are therefore differences from a reference that
  is biased downward toward leakier estimates on non-iid data, and
  conservative lower bounds in that direction. The implication for
  cross-class magnitude \emph{ranking} (that Class I \textless{} Class
  II \textless{} Class III in prevalence-weighted impact) holds only
  under the assumption that the baseline absorbs each class's
  contamination proportionally. That assumption is consistent with the
  within-subject design (every class is measured against the same clean
  baseline per dataset) but is not separately verified by an oracle
  comparison. A future replication against a population-level oracle on
  a synthetic data generator would isolate the baseline component from
  the leakage signal.
\item
  \textbf{Noise/diversity decomposition rests on a zero-diversity
  assumption for seeds.} The decomposition
  \(\Delta = \Delta_{\text{noise}} + \Delta_{\text{diversity}}\) uses
  seed inflation as a direct estimate of the noise component, justified
  by the modelling assumption that different seeds estimate the same
  true AUC (zero diversity). The assumption is plausible (same
  algorithm, same data, only the initialization differs), but the paper
  does not independently measure inter-seed prediction correlation per
  dataset to verify that seed-induced AUC variation is exclusively
  sampling noise rather than partly reflecting genuinely different
  learned models. The 90\%/10\% split at the corpus median should
  therefore be read as consistent with the assumed zero-diversity
  decomposition rather than as an independently measured quantity.
  Direct measurement of inter-seed prediction correlation per dataset is
  the natural next test.
\item
  \textbf{Meta-regression class/algorithm confound.} In the
  Section~\ref{sec-meta-regression} hierarchical model, ``experiment''
  maps one-to-one to leakage class, and certain algorithms appear only
  in one class (DT in Class III, NB largely in Class III). The
  between-class variance estimate therefore partially reflects
  between-algorithm variance for those experiments. The qualitative
  conclusion that mechanism is the dominant moderator survives in
  Section~\ref{sec-raw-correlations} (where Class II/III effects are an
  order of magnitude above the d\(_z\) detection floor across all
  algorithm subsets), but the meta-regression's variance-ratio
  quantification is not an algorithm-free contrast.
\item
  \textbf{Corpus selection.} The 2,047-dataset corpus is dominated by
  OpenML benchmarks (clean iid tabular academic data), supplemented by
  PMLB (curated benchmark collection) and the \texttt{ml} package
  (datasets from prior work by the same author). The corpus does not
  represent modern high-stakes domains: medical records, financial time
  series, production ML logs.
\end{enumerate}

\section{Conclusion}\label{conclusion}

Data leakage is widely acknowledged as a threat to machine learning
validity, yet to my knowledge no prior study has quantified how much
each type of leakage actually costs. I present the first large-scale
quantitative landscape, measuring twenty-eight core leakage experiments
plus a boundary experiment across 2,047 benchmark datasets with internal
validation (key effects replicate on the confirmation dataset split and
out-of-fold holdout sets) and directional prediction tracking (10/13
confirmed, 2 falsified, 1 qualified; the failures are mechanistically
informative).

The central finding is a categorical distinction between four leakage
classes organized by causal mechanism:

\textbf{Estimation leakage} (the most commonly taught and most commonly
worried about) produces near-zero effects at typical dataset sizes on
iid tabular data (\(\Delta\)AUC \textless{} 0.005). Nine experiments
confirm this. The correct workflow practice (per-fold preprocessing)
should still be followed (scikit-learn Pipelines enforce it at zero
cost), but the pedagogical emphasis on this error type is
disproportionate to its measured effect.

\textbf{Selection leakage} (using test-set performance to guide model
selection, seed choice, or hyperparameter tuning) produces the dominant
effects at practical dataset sizes (\(\Delta\)AUC = +0.013 to +0.045,
d\(_z\) = 0.27--0.93). Every selection mechanism decomposes into noise
exploitation (decaying as \(1/\sqrt{n}\)) and genuine diversity (the
true performance spread across the selection pool). At the corpus median
(\(n \approx 1{,}900\)), the data are consistent with approximately 90\%
of the measured effect being noise exploitation under the
seed-as-zero-diversity assumption (Limitation 11). Seed inflation
measures that noise and vanishes by \(n = 5{,}000\); peeking retains a
residual (approximately +0.03 AUC at \(n = 100{,}000\)) that reflects
genuine algorithm diversity, not persistent leakage.

\textbf{Memorization leakage} (training on duplicated evaluation data)
produces effects amplified by model capacity, with a monotonic capacity
ordering NB (\(d_z =\) 0.37) \(<\) LR \(<\) XGB \(<\) RF \(<\) KNN \(<\)
DT (\(d_z =\) 1.11) at 10\% duplication. SMOTE and random oversampling
produce indistinguishable leakage (matched on mean, SD, IQR, skew, and
kurtosis).

\textbf{Class IV leakage} is invisible under random cross-validation
because random folding destroys the structure that would reveal it
(\citeproc{ref-roberts2017cross}{Roberts et al. 2017}). On 14 datasets
with verified genuine timestamps, the pure temporal effect (after
controlling for training-set size) averages +0.023 AUC; 92 FOREX
datasets confirm the null at zero. The effect is domain-dependent: near
zero on typical benchmarks, substantial where real concept drift exists.

\textbf{Cross-validation standard errors} under the naive z-based
default underestimate true uncertainty by approximately 1.7 \(\times\),
achieving 55\% actual coverage at nominal 95\% confidence (Wilson 95\%
CI {[}53.8\%, 56.4\%{]}). Conservative-Z (fold SD without
\(\div\sqrt{k}\)) closes most of the gap to approximately 87.4\% and is
the recommended interval.

The primary evidence is the raw, algorithm-free pattern
(Table~\ref{tbl-raw-correlations}): Class II/III effects are an order of
magnitude above the detection floor across every algorithm subset, while
no dataset moderator (sample size, feature count, class imbalance)
tracks effect size at within-class scale. A Bayesian hierarchical
meta-regression (12,103 observations) is \emph{consistent with} this
(between-experiment variance exceeds between-dataset variance by
2.6\(\times\)), but the meta-regression ratio is partially confounded
with algorithm (Limitation 12).

Two additional findings extend the picture further. First, feature
selection leakage scales with dimensionality: negligible at typical
\(p/n\) ratios, but +0.018 mean at \(p/n >\) 0.1, confirming Ambroise
and McLachlan (\citeproc{ref-ambroise2002selection}{2002}) at scale.
Second, metric selection flips model rankings on 31\% of datasets; the
researcher who reports the most flattering metric is making a selection
decision as real as peeking at the test set.

If these magnitude rankings generalize beyond tabular binary
classification (an open question), the priority implication is
\textbf{prevalence-weighted, not magnitude-only}: fix selection leakage
first because peeking-style selection is universal across the iid
tabular regime at practical \(n\) (every model-selection decision
exposes the holdout regardless of dataset properties), though seed-style
selection decays to zero by \(n \geq 5{,}000\); audit for boundary
effects in temporal or high-dimensional domains second; audit for
memorization leakage in high-capacity models third (raw d\(_z\) for
Class III at high capacity exceeds Class II: DT at 30\% duplication
reaches d\(_z\) = 1.38 versus peeking d\(_z\) = 0.93, but Class III only
manifests in duplicate-prone or capacity-amplified contexts, so its
prevalence-weighted impact is lower); and deprioritize estimation
leakage on iid tabular data, with one carve-out: target encoding (Exp
AC) has Class-I mechanism but Class-II-magnitude inflation and must
still be done per-fold. The researcher who normalizes before splitting
has committed a textbook sin that costs on average nothing. The
researcher who peeks at the test set has no honest assessment left;
every decision built on that score is built on sand.

\section{Data and Code Availability}\label{data-and-code-availability}

Experiment runners, figure generation scripts, the Bayesian
meta-regression script, \texttt{claims.json} (the authoritative values
behind every interpolated quantity in the text), and
\texttt{citations.public.json} (paper-level citation provenance: prose
anchor + sha256 hash + bibkey list per sentence) are at
\href{https://github.com/epagogy/ml}{github.com/epagogy/ml}. Raw L0
JSONL result files are not yet public; their release is queued for the
next iteration.

\section*{Conflict of Interest}\label{conflict-of-interest}
\addcontentsline{toc}{section}{Conflict of Interest}

The author develops the \texttt{ml} software package, which implements
the grammar described in the companion paper
(\citeproc{ref-roth2026grammar}{Roth 2026}). The leakage experiments
reported here are independent of that software and use scikit-learn
throughout. The dataset corpus includes datasets sourced through ml
(\textless{} 5\% of the 2,047 total); excluding ml-sourced datasets does
not change any finding. The priority ordering reported in the Conclusion
(selection \textgreater{} boundary \textgreater{} memorization
\textgreater{} estimation) emerged from measured magnitudes in this
corpus and was not chosen to match the grammar's design; the grammar's
four constraints address all four leakage classes.

\section*{Reproducibility}\label{reproducibility}
\addcontentsline{toc}{section}{Reproducibility}

All experiments use Python 3.11, scikit-learn 1.4, XGBoost 2.0, NumPy
1.26, PyMC 5.28 with numpyro 0.14 backend. Random seeds are fixed at 42
for every CV split and stochastic algorithm initialization;
seed-inflation experiments (AI, AP) use consecutive integer seeds
starting at 0. The \texttt{claims.json} payload (a sibling of this
manuscript) holds the authoritative values backing every interpolated
quantity in this paper. A single-command reproducibility bundle that
re-runs every experiment from corpus snapshot to figure is in
development; it will support a planned replication study.

\section*{Acknowledgments}\label{acknowledgments}
\addcontentsline{toc}{section}{Acknowledgments}

This work was conducted independently and received no external funding.
I thank the colleagues and peers who provided critical feedback during
this process; they will be acknowledged individually upon journal
submission, if they choose to be named. This is ongoing work; feedback
is welcome at simon@epagogy.ai.

\section*{Disclosure}\label{disclosure}
\addcontentsline{toc}{section}{Disclosure}

Large language models (Claude, Anthropic) were used as principal
writing, analysis, and implementation tools during the preparation of
this manuscript. All scientific claims, experimental designs, empirical
results, and theoretical contributions are my own. I take full
responsibility for the content.

\section*{References}\label{references}
\addcontentsline{toc}{section}{References}

\phantomsection\label{refs}

\end{document}